\def\year{2015}
\documentclass[letterpaper]{article}
\usepackage{aaai}
\usepackage{times}
\usepackage{helvet}
\usepackage{courier}
\usepackage{amsmath}
\usepackage{amssymb}
\usepackage{graphicx}
\usepackage{floatrow}

\newtheorem{defi}{Definition}
\newtheorem{theorem}{Theorem}
\newtheorem{lemma}{Lemma}
\newtheorem{proposition}{Proposition}

\def\myproof{1} 
\begin{document}
\newcommand{\set}[1]{{\mathcal #1}}
\newcommand{\sset}[1]{{|\mathcal #1|}}

\title{Parallel Gaussian Process Regression for Big Data:\\ Low-Rank Representation Meets Markov Approximation}

\author{
Kian Hsiang Low\thanks{\protect Kian Hsiang Low and Jiangbo Yu are co-first authors.} \and Jiangbo Yu$^{\ast}$ \and Jie Chen$^{\S}$ \and Patrick Jaillet$^{\dag}$\\
Department of Computer Science, National University of Singapore, Republic of Singapore$^{\ast}$\\
Singapore-MIT Alliance for Research and Technology, Republic of Singapore$^{\S}$\\
Massachusetts Institute of Technology, USA$^{\dag}$\\
\{lowkh, yujiang\}@comp.nus.edu.sg$^{\ast}$, chenjie@smart.mit.edu$^{\S}$, jaillet@mit.edu$^{\dag}$
}
\maketitle

\begin{abstract}
\begin{quote}
The expressive power of a \emph{Gaussian process} (GP) model comes at a cost of poor scalability in the data size.
To improve its scalability, this paper presents a \emph{\underline{l}ow-rank-cum-\underline{M}arkov \underline{a}pproximation} (LMA) of the GP model that is novel in leveraging the dual computational advantages stemming from complementing a low-rank approximate representation of the full-rank GP based on a support set of inputs with a Markov approximation of the resulting residual process; the latter approximation is guaranteed to be closest in the Kullback-Leibler distance criterion subject to some constraint
and is considerably more refined than that of existing sparse GP models utilizing low-rank representations due to its more relaxed conditional independence assumption (especially with larger data).
As a result, our LMA method can trade off between the size of the support set and the order of the Markov property to (a) incur lower computational cost than such sparse GP models while achieving predictive performance comparable to them and (b) accurately represent features/patterns of any scale.
Interestingly, varying the Markov order produces a spectrum of LMAs
with PIC approximation and full-rank GP at the two extremes.
An advantage of our LMA method is that it is amenable to parallelization on multiple machines/cores, thereby gaining greater scalability.
Empirical evaluation on three real-world datasets in clusters of up to $32$ computing nodes shows that our centralized and parallel LMA methods are significantly more time-efficient and scalable than state-of-the-art sparse and full-rank GP regression methods 
while achieving comparable predictive performances.
\end{quote}
\end{abstract}

\section{Introduction}
\label{sec:intro}
%
\emph{Gaussian process} (GP) models are a rich class of Bayesian non-parametric models that can perform probabilistic regression by providing Gaussian predictive distributions with formal measures of the predictive uncertainty.
Unfortunately, a GP model is handicapped by its poor scalability in the size of the data, hence limiting its practical use to small data.
To improve its scalability, two families of sparse GP regression methods have been proposed: 
(a) Low-rank approximate representations \cite{Lawrence13,candela10,candela05,snelson06} of the \emph{full-rank GP} (FGP) model are well-suited for modeling slowly-varying functions with large correlation
and can use all the data for predictions. But, they require a relatively high rank to capture small-scale features/patterns (i.e., of small correlation) with high fidelity, thus losing their computational advantage.
(b) In contrast, localized regression and covariance tapering methods (e.g., local GPs \cite{Park11} and compactly supported covariance functions \cite{Furrer06}) are particularly useful for modeling rapidly-varying functions with small correlation. However, they can only utilize local data for predictions, thereby performing poorly in input regions with little/no data. Furthermore, to accurately represent large-scale features/patterns (i.e., of large correlation), the locality/tapering range has to be increased considerably, thus sacrificing their time efficiency.

Recent sparse GP regression methods \cite{LowUAI13,snelson07} have unified approaches from the two families described above to harness their complementary modeling and predictive capabilities (hence, eliminating their deficiencies) while retaining their computational advantages.
Specifically, after approximating the FGP (in particular, its covariance matrix) with a low-rank representation based on the notion of a support set of inputs, a sparse covariance matrix approximation of the resulting residual process is made. 
However, this sparse residual covariance matrix approximation imposes a fairly strong conditional independence assumption given the support set since the support set cannot be too large to preserve time efficiency (see Remark $2$ after Proposition~\ref{prop1} in Section~\ref{sgprlrma}).
In this paper, we argue that such a strong assumption is an overkill: It is in fact possible to construct a more refined, \emph{dense} residual covariance matrix approximation by exploiting a Markov assumption and, perhaps surprisingly, still achieve scalability, which distinguishes our work here from existing sparse GP regression methods utilizing low-rank representations (i.e., including the unified approaches) described earlier.
As a result, our proposed residual covariance matrix approximation can significantly relax the conditional independence assumption (especially with larger data; see Remark $1$ after Proposition~\ref{prop1} in Section~\ref{sgprlrma}), hence potentially improving the predictive performance.

This paper presents a \emph{\underline{l}ow-rank-cum-\underline{M}arkov \underline{a}pproximation} (LMA) of the FGP model (Section~\ref{sgprlrma}) that is novel in leveraging the dual computational advantages stemming from complementing the reduced-rank covariance matrix approximation based on the support set with the residual covariance matrix approximation due to the Markov assumption;
the latter approximation is guaranteed to be closest in the Kullback-Leibler distance criterion subject to some constraint.
Consequently, our proposed LMA method can trade off between the size of the support set and the order of the Markov property to (a) incur lower computational cost than sparse GP regression methods utilizing low-rank representations with only the 
support set size (e.g., \cite{LowUAI13,snelson07}) or number of spectral points \cite{candela10}
as the varying parameter while achieving predictive performance comparable to them and (b) accurately represent features/patterns of any scale.
Interestingly, varying the Markov order produces a spectrum of LMAs
with the \emph{partially independent conditional} (PIC) approximation \cite{LowUAI13,snelson07} and FGP at the two extremes.
An important advantage of LMA over most existing sparse GP regression methods is that it is amenable to parallelization on multiple machines/cores, thus gaining greater scalability for performing real-time predictions necessary in many time-critical applications and decision support systems (e.g., ocean sensing \cite{LowAAMAS13,LowSPIE09,LowAAMAS08,low09,LowAAMAS11,LowAAMAS12,LowAeroconf10}, traffic monitoring \cite{chen12,LowRSS13,LowECML14b,NghiaICML14,LowDyDESS14,LowECML14a,LowAAMAS14,LowAAAI14,LowIAT12}).
Our parallel LMA method is implemented using the \emph{message passing interface} (MPI) framework to run in clusters of up to $32$ computing nodes and its predictive performance, scalability, and speedup are empirically evaluated on three real-world datasets (Section~\ref{expts}).
\section{Full-Rank Gaussian Process Regression}
\label{sect:gpr}
%
Let $\set{X}$ be a set representing the input domain such that each input $x\in \set{X}$ denotes a $d$-dimensional feature vector and is associated with a realized output value $y_x$ (random output variable $Y_x$) if it is observed (unobserved). 
Let $\{Y_x\}_{x \in \set{X}}$ denote a GP, that is, every finite subset of $\{Y_x\}_{x \in \set{X}}$ follows a multivariate Gaussian distribution.
Then, the GP is fully specified by its \emph{prior} mean $\mu_{x} \triangleq \mathbb{E}[Y_x]$ and covariance $\sigma_{x x'} \triangleq \mbox{cov}[Y_x, Y_{x'}]$ for all $x, x' \in \set{X}$.
Supposing a column vector $y_\set{D}$ of realized outputs is observed for some set $\set{D}\subset\set{X}$ of inputs,
a \emph{full-rank GP} (FGP) model 
can perform probabilistic regression
by providing a Gaussian \emph{posterior}/predictive distribution
$$\set{N}(\mu_\set{U} + \Sigma_{\set{U}\set{D}}\Sigma_{\set{D}\set{D}}^{-1}( y_\set{D} - \mu_\set{D} ), \Sigma_{\set{U}\set{U}}-\Sigma_{\set{U}\set{D}}\Sigma_{\set{D}\set{D}}^{-1}\Sigma_{\set{D}\set{U}})$$
of the unobserved outputs for any set $\set{U}\subseteq \set{X}\setminus \set{D}$ of inputs 
%
%
where ${\mu}_\set{U}$ (${\mu}_\set{D}$) is a column vector with mean components $\mu_{x}$ for all $x\in \set{U}$ ($x\in \set{D}$),
$\Sigma_{\set{U}\set{D}}$ ($\Sigma_{\set{D}\set{D}}$) is a covariance matrix with covariance components $\sigma_{xx'}$ for all $x\in \set{U}, x'\in \set{D}$ ($x, x'\in \set{D}$), and $\Sigma_{\set{D}\set{U}}=\Sigma^{\top}_{\set{U}\set{D}}$. 
The chief limitation hindering the practical use of the FGP regression method 
is its poor scalability in the data size $|\set{D}|$: 
Computing the Gaussian posterior/predictive distribution
requires inverting $\Sigma_{\set{D}\set{D}}$, which incurs $\set{O}(|\set{D}|^3)$ time.
In the next section, we will introduce our proposed LMA method to improve its scalability.
\section{Low-Rank-cum-Markov Approximation}
\label{sgprlrma}
%
%
$\widehat{Y}_x\triangleq\Sigma_{x\set{S}}\Sigma^{-1}_{\set{S}\set{S}}Y_{\set{S}}$ is a reduced-rank approximate representation of $Y_x$ based on a \emph{support set} $\set{S}\subset\set{X}\vspace{-0mm}$ of inputs and its finite-rank covariance function is
$\mbox{cov}[\widehat{Y}_x, \widehat{Y}_{x'}]\vspace{-0mm}=\Sigma_{x\set{S}}\Sigma^{-1}_{\set{S}\set{S}}\Sigma_{\set{S}x'}$
for all $x,x'\in\set{X}$.
Then, $\widetilde{Y}_x = Y_x - \widehat{Y}_x$ is the residual of the reduced-rank approximation\vspace{-0mm} and its covariance function is thus 
$\mbox{cov}[\widetilde{Y}_x, \widetilde{Y}_{x'}]\vspace{-0mm}=\sigma_{xx'}-\Sigma_{x\set{S}}\Sigma^{-1}_{\set{S}\set{S}}\Sigma_{\set{S}x'}$.
Define
$$Q_{\set{B}\set{B}'}\triangleq\Sigma_{\set{B}\set{S}}\Sigma^{-1}_{\set{S}\set{S}}\Sigma_{\set{S}\set{B'}}\ \ \ \text{and}\ \ \ R_{\set{B}\set{B}'}\triangleq\Sigma_{\set{B}\set{B}'}-Q_{\set{B}\set{B}'}\vspace{-0mm}$$
for all $\set{B}, \set{B}'\subset\set{X}$. 
Then, a covariance matrix $\Sigma_{\set{V}\set{V}}$ for the set $\set{V}\triangleq\set{D}\cup\set{U}\subset\set{X}$ of inputs (i.e., associated with realized outputs $y_\set{D}$ and unobserved random outputs $Y_\set{U}$) can be decomposed into a reduced-rank covariance matrix approximation $Q_{\set{V}\set{V}}$ 
and the resulting residual covariance matrix $R_{\set{V}\set{V}}$,
that is, $\Sigma_{\set{V}\set{V}} = Q_{\set{V}\set{V}}+ R_{\set{V}\set{V}}$.
As discussed in Section~\ref{sec:intro},
existing sparse GP regression methods utilizing low-rank representations (i.e., including unified approaches) 
approximate $R_{\set{V}\set{V}}$ with a sparse matrix.
In contrast, we will construct a more refined, dense residual covariance matrix approximation by exploiting a Markov assumption to be described next.

Let the set $\set{D}$ ($\set{U}\vspace{-0mm}$) of inputs 
be partitioned\footnote{$\set{D}$ and $\set{U}$ are partitioned according to a simple parallelized clustering scheme employed in the work of \citeauthor{LowUAI13}~\shortcite{LowUAI13}.} evenly into $M$ disjoint subsets $\set{D}_1,\ldots,\set{D}_M$ ($\set{U}_1,\ldots,\set{U}_M$)
such that the outputs $y_{\set{D}_m}$ and $Y_{\set{U}_m}\vspace{-0mm}$ are as highly correlated as possible for $m=1,\ldots,M$.
Let $\set{V}_m\triangleq\set{D}_m\cup\set{U}_m$.
Then, $\set{V}=\bigcup^{M}_{m=1}\set{V}_m$.

The key idea of our \emph{\underline{l}ow-rank-cum-\underline{M}arkov \underline{a}pproximation} (LMA) method is to approximate the residual covariance matrix $R_{\set{V}\set{V}}$ by a block matrix $\overline{R}_{\set{V}\set{V}}$ partitioned into $M\times M$ square blocks, that is,
$
\overline{R}_{\set{V}\set{V}}\triangleq[\overline{R}_{\set{V}_m\set{V}_{n}}]_{m,n=1,\ldots,M}
$
where\vspace{-0mm}
\begin{equation}
\displaystyle\overline{R}_{\set{V}_m\set{V}_{n}}
\hspace{-0.5mm}\triangleq\hspace{-0.5mm}\left\{ \hspace{-1mm}
\begin{array}{ll}
{R}_{\set{V}_m\set{V}_{n}} &\hspace{-1mm} \text{if}\ |m-n|\leq B,\vspace{0mm}\\
R_{\set{V}_m\set{D}^B_{m}}R_{\set{D}^B_{m}\set{D}^B_{m}}^{-1} \overline{R}_{\set{D}^B_{m}\set{V}_{n}} &\hspace{-1mm}  \text{if}\ n-m>B>0,\vspace{0mm}\\
\overline{R}_{\set{V}_{m}\set{D}^B_{n}}R_{\set{D}^B_{n}\set{D}^B_{n}}^{-1} R_{\set{D}^B_{n}\set{V}_{n}} &\hspace{-1mm}  \text{if}\ m-n>B>0,\vspace{0mm}\\
\underline{0} & \hspace{-1mm}\text{if}\ |m-n| > B =0;\vspace{-0mm}
\end{array}\right. \vspace{-0mm}
\label{ares}
\end{equation}
such that $B\in\{0,\ldots,M-1\}\vspace{-0mm}$ denotes the order of the Markov property imposed on the residual process $\{\widetilde{Y}_x\}_{x\in\set{D}}$ 
to be detailed later,
$\set{D}^B_{m} \triangleq\bigcup^{\min(m+B,M)}_{k=m+1}\set{D}_{k}$,
and $\underline{0}$ denotes a 
square block comprising components of value $0$.

\begin{figure}
\begin{tabular}{cc}
\includegraphics[scale=0.2]{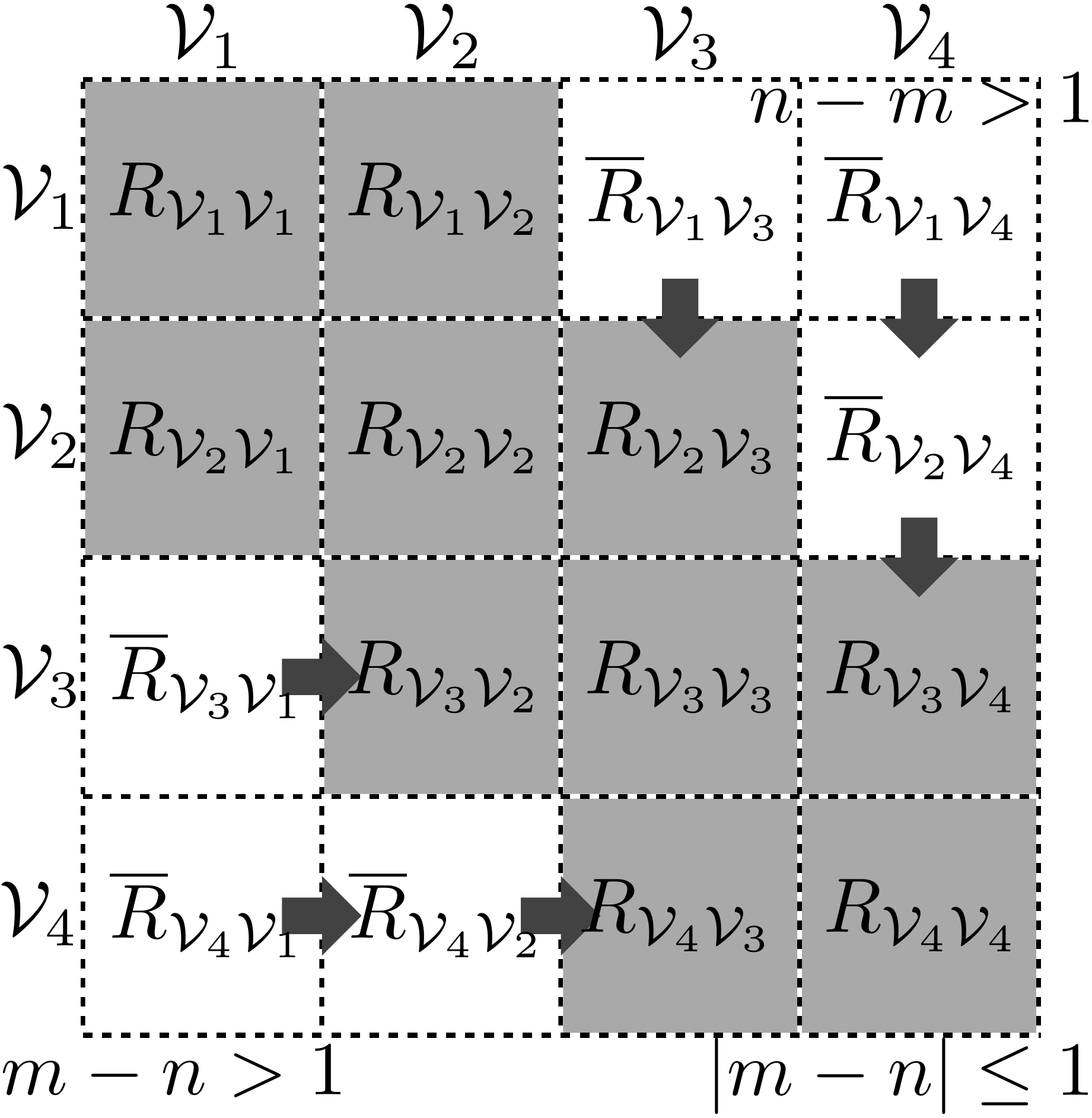} & \includegraphics[scale=0.2]{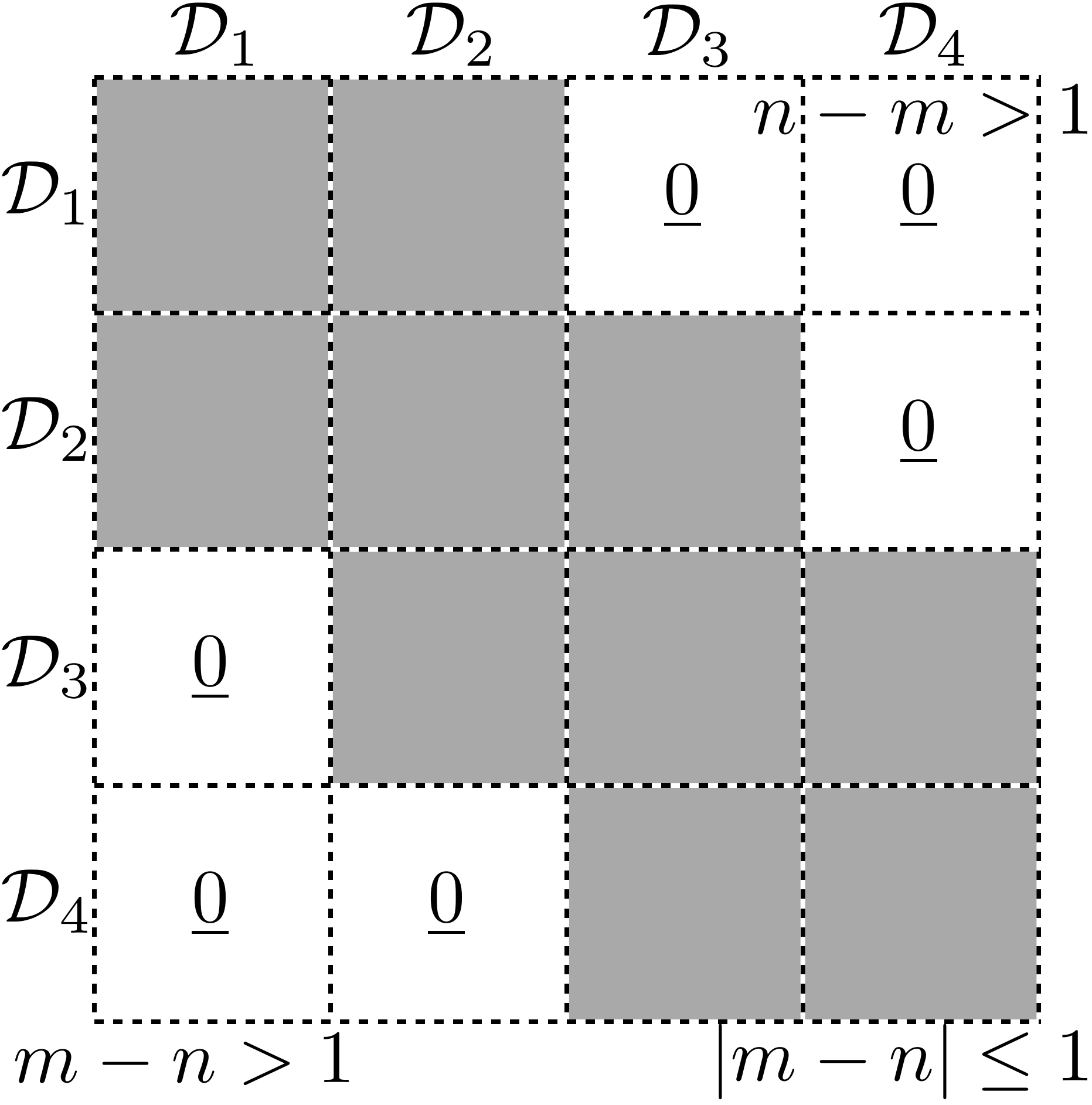}\vspace{-1.5mm}\\
(a) $\overline{R}_{\set{V}\set{V}}$ & (b) $\overline{R}^{-1}_{\set{D}\set{D}}$\vspace{-3.5mm}
\end{tabular}
\caption{$\overline{R}_{\set{V}\set{V}}$ and $\overline{R}^{-1}_{\set{D}\set{D}}$ with $B=1$ and $M=4$. (a) Shaded blocks (i.e., $|m-n|\leq B$) form the $B$-block band while unshaded blocks (i.e., $|m-n|>B$) fall outside the band. Each arrow denotes a recursive call. (b) Unshaded blocks\vspace{-0mm} outside $B$-block band of $\overline{R}^{-1}_{\set{D}\set{D}}$ (i.e., $|m-n|> B$) are $\underline{0}$.\vspace{-4mm}}
\label{fig:matrix}
\end{figure}

To understand the intuition underlying the approximation in \eqref{ares}, 
Fig.~\ref{fig:matrix}a illustrates a simple case of $\overline{R}_{\set{V}\set{V}}$ with $B=1$ and $M=4$ for ease of exposition: It can be observed that only the blocks ${R}_{\set{V}_m\set{V}_{n}}$ \emph{outside} the $B$-block band of $R_{\set{V}\set{V}}$ (i.e., $|m-n|>B$) are approximated, specifically, by unshaded blocks $\overline{R}_{\set{V}_m\set{V}_{n}}$ being defined 
as a recursive series of $|m-n|-B$ reduced-rank residual covariance matrix approximations \eqref{ares}.
So, when an unshaded block $\overline{R}_{\set{V}_m\set{V}_{n}}$ is further from the diagonal of $\overline{R}_{\set{V}\set{V}}$ (i.e., larger $|m-n|$), it is derived using more reduced-rank residual covariance matrix approximations. 
For example, ${R}_{\set{V}_1\set{V}_{4}}$ is approximated by an unshaded block $\overline{R}_{\set{V}_1\set{V}_{4}}$ being defined as a recursive series of $2$ reduced-rank residual covariance matrix approximations, namely, 
approximating ${R}_{\set{V}_1\set{V}_{4}}\vspace{-0mm}$ by
$R_{\set{V}_1\set{D}^1_{1}}R_{\set{D}^1_{1}\set{D}^1_{1}}^{-1} {R}_{\set{D}^1_{1}\set{V}_{4}}=R_{\set{V}_1\set{D}_{2}}R_{\set{D}_{2}\set{D}_{2}}^{-1} {R}_{\set{D}_{2}\set{V}_{4}}\vspace{-0mm}$ based on the support set $\set{D}^1_{1}=\set{D}_{2}$ of inputs and in turn approximating ${R}_{\set{D}_{2}\set{V}_{4}}$ by a submatrix $\overline{R}_{\set{D}_{2}\set{V}_{4}} = R_{\set{D}_{2}\set{D}_{3}}R_{\set{D}_{3}\set{D}_{3}}^{-1} {R}_{\set{D}_{3}\set{V}_{4}}\vspace{-0mm}$ \eqref{ares} of unshaded block $\overline{R}_{\set{V}_{2}\set{V}_{4}}$ based on the support set $\set{D}^1_{2}=\set{D}_{3}\vspace{-0mm}$ of inputs.
As a result, $\overline{R}_{\set{V}_1\set{V}_{4}}=R_{\set{V}_1\set{D}_{2}}R_{\set{D}_{2}\set{D}_{2}}^{-1} R_{\set{D}_{2}\set{D}_{3}}R_{\set{D}_{3}\set{D}_{3}}^{-1} {R}_{\set{D}_{3}\set{V}_{4}}$ is fully specified by five submatrices 
of the respective shaded blocks $R_{\set{V}_1\set{V}_{2}}$, $R_{\set{V}_{2}\set{V}_{2}}$, $R_{\set{V}_{2}\set{V}_{3}}$, $R_{\set{V}_{3}\set{V}_{3}}$, and ${R}_{\set{V}_{3}\set{V}_{4}}$ \emph{within} the $B$-block band of $\overline{R}_{\set{V}\set{V}}$ (i.e., $|m-n|\leq B$).
In general, any unshaded block
$\overline{R}_{\set{V}_{m}\set{V}_{n}}$ 
\emph{outside} the $B$-block band of $\overline{R}_{\set{V}\set{V}}$ (i.e., $|m-n|>B$) is fully specified by submatrices of the shaded blocks \emph{within} the $B$-block band of $\overline{R}_{\set{V}\set{V}}$ (i.e., $|m-n|\leq B$) due to its recursive series of $|m-n|-B$ reduced-rank residual covariance matrix approximations \eqref{ares}.
%
%
Though it may not be obvious now how such an approximation would entail scalability, \eqref{ares} interestingly offers an alternative interpretation of imposing a $B\vspace{-0mm}$-th order Markov property on residual process $\{\widetilde{Y}_x\}_{x\in\set{D}}$, which reveals a further insight on the structural assumption of LMA to be exploited for achieving scalability, as detailed later.

The covariance matrix $\Sigma_{\set{V}\set{V}}$ is thus approximated by a block matrix $\overline{\Sigma}_{\set{V}\set{V}}\triangleq Q_{\set{V}\set{V}} + \overline{R}_{\set{V}\set{V}}$ partitioned into $M\times M$ square blocks, that is, 
$
\overline{\Sigma}_{\set{V}\set{V}}\triangleq[\overline{\Sigma}_{\set{V}_m\set{V}_{n}}]_{m,n=1,\ldots,M}
$
where\vspace{-0mm}
\begin{equation}
\hspace{-0mm}
\displaystyle\overline{\Sigma}_{\set{V}_m\set{V}_{n}}
\triangleq
Q_{\set{V}_m\set{V}_n} +\overline{R}_{\set{V}_m\set{V}_{n}}\ . 
\label{acov}\vspace{-0mm}
\end{equation}
So, within the $B$-block band of $\vspace{-0mm}\overline{\Sigma}_{\set{V}\set{V}}$ (i.e., $|m-n|\leq B$), $\overline{\Sigma}_{\set{V}_m\set{V}_{n}}={\Sigma}_{\set{V}_m\set{V}_{n}}$, by \eqref{ares} and \eqref{acov}.
Note that when $B=0$, $\overline{\Sigma}_{\set{V}_m\set{V}_{n}}=Q_{\set{V}_m\set{V}_n}$ for $|m-n|>B$, thus yielding the prior covariance matrix $\overline{\Sigma}_{\set{V}\set{V}}$ of the \emph{partially independent conditional} (PIC) approximation \cite{LowUAI13,snelson07}.
When $B=M-1$, $\overline{\Sigma}_{\set{V}\set{V}} = \Sigma_{\set{V}\set{V}}$ is the prior covariance matrix of FGP model.
So, LMA generalizes PIC (i.e., if $B=0$) and
becomes FGP if $B=M-1$.
Varying Markov order $B$ from $0$ to $M-1$ produces a spectrum of LMAs with PIC and FGP at the two extremes.

By approximating $\Sigma_{\set{V}\set{V}}$ with $\overline{\Sigma}_{\set{V}\set{V}}$,
our LMA method utilizes the data $(\set{D},y_{\set{D}})$ to predict the unobserved outputs
for any set $\set{U}\subseteq \set{X}\setminus \set{D}$ of inputs and provide their corresponding predictive uncertainties
using the following predictive mean vector and covariance matrix, respectively:\vspace{-1.5mm}
\begin{equation}
{\mu}^{\text{\tiny{LMA}}}_{{\set{U}}}\triangleq\displaystyle\mu_{\set{U}}+\overline{\Sigma}_{{\set{U}}\set{D}}\overline{\Sigma}_{\set{D}\set{D}}^{-1}\left(y_\set{D}-\mu_\set{D} \right) \vspace{-1.1mm}
\label{vapmu}
\end{equation}
\begin{equation}
{\Sigma}^{\text{\tiny{LMA}}}_{{\set{U}}{\set{U}}}
\displaystyle\triangleq\overline{\Sigma}_{\set{U}{\set{U}}}-\overline{\Sigma}_{{\set{U}}\set{D}}\overline{\Sigma}_{\set{D}\set{D}}^{-1}\overline{\Sigma}_{\set{D}{\set{U}}}\vspace{-0mm}
\label{vapvar}
\end{equation}
where $\vspace{-0mm}\overline{\Sigma}_{{\set{U}}\set{U}}$, $\overline{\Sigma}_{{\set{U}}\set{D}}$, and $\overline{\Sigma}_{\set{D}\set{D}}$ are obtained using \eqref{acov}, and $\overline{\Sigma}_{\set{D}{\set{U}}}=\overline{\Sigma}^{\top}_{{\set{U}}\set{D}}$.
If $\overline{\Sigma}_{\set{D}\set{D}}$ in \eqref{vapmu} and \eqref{vapvar} is inverted directly,
then it would still incur the same $\set{O}(|\set{D}|^3)$ time as inverting ${\Sigma}_{\set{D}\set{D}}$ in 
the FGP regression method (Section~\ref{sect:gpr}).
In the rest of this section, we will show how this scalability issue can be resolved by leveraging the computational advantages associated with both the reduced-rank covariance matrix approximation $Q_{\set{D}\set{D}}$ based on the support set $\set{S}$ and our proposed residual covariance matrix approximation $\overline{R}_{\set{D}\set{D}}$ 
due to $B$-th order Markov assumption after decomposing $\overline{\Sigma}_{\set{D}\set{D}}$.

It can be observed from $\overline{R}_{\set{V}\set{V}}$ \eqref{ares} that
$R_{\set{D}\set{D}}$ 
is approximated by a block matrix 
$
\overline{R}_{\set{D}\set{D}}=[\overline{R}_{\set{D}_m\set{D}_{n}}]_{m,n=1,\ldots,M}
$
where $\overline{R}_{\set{D}_m\set{D}_{n}}$ is a submatrix of $\overline{R}_{\set{V}_m\set{V}_{n}}$ obtained using \eqref{ares}.
\begin{proposition}
Block matrix $\overline{R}^{-1}_{\set{D}\set{D}}$ is $B$-block-banded, that is, 
any block outside its $B$-block band is $\underline{0}$ (e.g., Fig.~\ref{fig:matrix}b).
\label{prop1}
\end{proposition}
Its proof follows directly from a block-banded matrix result of \citeauthor{Moura05}~\shortcite{Moura05} (specifically, Theorem $3$).\vspace{0.5mm}


\noindent
\textit{Remark} $1$. In the same spirit as a Gaussian Markov random process, imposing a $B$-th order Markov property on residual process $\{\widetilde{Y}_x\}_{x\in\set{D}}$ is equivalent to approximating $R_{\set{D}\set{D}}$ by $\overline{R}_{\set{D}\set{D}}$ whose inverse is $B$-block-banded (Fig.~\ref{fig:matrix}b).
That is, if $|m-n|> B$, $Y_{\set{D}_m}$ and $Y_{\set{D}_n}$ are conditionally independent given $Y_{\set{S}\cup\set{D}\setminus(\set{D}_m\cup\set{D}_n)}$.
Such a conditional independence assumption thus becomes more relaxed with larger data.
More importantly, this $B\vspace{-0mm}$-th order Markov assumption or, equivalently, sparsity of $B$-block-banded $\overline{R}^{-1}_{\set{D}\set{D}}$ is the key to achieving scalability, as shown in the proof of Theorem~\ref{thm2} later.

\noindent
\textit{Remark} $2$. Though $\overline{R}^{-1}_{\set{D}\set{D}}$ is sparse, $\overline{R}_{\set{D}\set{D}}$ is a \emph{dense} residual covariance matrix approximation if $B>0$. In contrast, the sparse GP regression methods utilizing low-rank representations (i.e., including unified approaches) 
utilize a sparse residual covariance matrix approximation (Section~\ref{sec:intro}), hence imposing a significantly stronger conditional independence assumption than LMA.
For example, PIC \cite{LowUAI13,snelson07} assumes $Y_{\set{D}_m}$ and $Y_{\set{D}_n}$ to be conditionally independent given only $Y_{\set{S}}$ if $|m-n|> 0$.\vspace{0.5mm}

The next result reveals that, among all $|\set{D}|\times |\set{D}|\vspace{-0mm}$ matrices whose inverse is $B$-block-banded, $\overline{R}_{\set{D}\set{D}}$ approximates ${R}_{\set{D}\set{D}}$ most closely in the Kullback-Leibler (KL) distance criterion, that is, $\overline{R}_{\set{D}\set{D}}$ has the minimum KL distance from ${R}_{\set{D}\set{D}}$:\vspace{-0mm}
\begin{theorem}
Let KL distance $\vspace{-0mm}D_\text{\em KL}(R,\widehat{R})\triangleq 0.5(\text{\em tr}(R\widehat{R}^{-1})-\log|R\widehat{R}^{-1}|-|\set{D}|)$ between two $|\set{D}|\times |\set{D}|$ positive definite matrices $R$ and $\widehat{R}\vspace{-0mm}$ 
measure the error of approximating ${R}$ with $\widehat{R}$.
Then, for any  matrix $\widehat{R}\vspace{-0mm}$ whose inverse is $B$-block-banded,  $D_\text{\em KL}({R}_{\set{D}\set{D}},\widehat{R})\geq D_\text{\em KL}({R}_{\set{D}\set{D}},\overline{R}_{\set{D}\set{D}})$.
\label{thm1}
\end{theorem}
Its proof is in\if\myproof1 Appendix~\ref{pthm1}. \fi\if\myproof0 \cite{AA14}. \fi
\vspace{-0mm} 
Our main result in Theorem~\ref{thm2} below exploits the sparsity of $\overline{R}^{-1}_{\set{D}\set{D}}$ (Proposition~\ref{prop1}) for
deriving an efficient formulation of
LMA, which is amenable to parallelization on multiple machines/cores by constructing and communicating the following summary information:
%
%
%
\begin{defi}[Local Summary] The $m$-th local summary is defined as a tuple $(\dot{y}_{m},\dot{R}_m, \dot{\Sigma}^{m}_{\set{S}},\dot{\Sigma}^{m}_{\set{U}})$  where\vspace{-0mm}
$$
\begin{array}{rl}
\displaystyle\dot{y}_{m}\triangleq &\hspace{-2mm} y_{\set{D}_m}\hspace{-0mm}-\hspace{-0mm}\mu_{\set{D}_m}  \hspace{-0mm} -\hspace{-0mm} R'_{\set{D}_m\set{D}^B_{m}}\hspace{-0mm}(y_{\set{D}^B_{m}}\hspace{-0mm}-\hspace{-0mm}\mu_{\set{D}^B_{m}})\vspace{0.5mm}\\
%
\displaystyle
\dot{R}_m \triangleq &\hspace{-2mm} (R_{\set{D}_m\set{D}_m} - R'_{\set{D}_m\set{D}^B_{m}} R_{\set{D}^B_{m}\set{D}_m})^{-1}\vspace{0.5mm}\\
\displaystyle\dot{\Sigma}^{m}_{\set{S}}\triangleq &\hspace{-2mm}\Sigma_{\set{D}_{m}\set{S}}\hspace{-0mm} -\hspace{-0mm} R'_{\set{D}_m\set{D}^B_{m}}\Sigma_{\set{D}^B_{m}\set{S}}\vspace{0.5mm}\\ 
\displaystyle\dot{\Sigma}^{m}_{\set{U}}\triangleq &\hspace{-2mm} \overline{\Sigma}_{\set{D}_{m}\set{U}} \hspace{-0mm}-\hspace{-0mm} 
R'_{\set{D}_m\set{D}^B_{m}}\overline{\Sigma}_{\set{D}^B_{m}\set{U}}\vspace{-0mm}
\end{array}
$$
%
such that $R'_{\set{D}_m\set{D}^B_{m}}\triangleq R_{\set{D}_m\set{D}^B_{m}} R_{\set{D}^B_{m}\set{D}^B_{m}}^{-1}$.
\label{def:ls}
\end{defi}
\begin{defi}[Global Summary] The global summary is defined as a tuple $(\ddot{y}_{\set{S}},\ddot{y}_{\set{U}},\ddot{\Sigma}_{\set{S}\set{S}},\ddot{\Sigma}_{\set{U}\set{S}},\ddot{\Sigma}_{\set{U}\set{U}})$ where\vspace{-0mm}
$$
\hspace{-1mm}
\begin{array}{rl}
\ddot{y}_{\set{S}}\hspace{-0mm}\triangleq &\hspace{-2mm}
\displaystyle\sum^M_{m=1}\hspace{-0mm} (\dot{\Sigma}^{m}_{\set{S}})^{\top} \hspace{-0mm}\dot{R}_m \dot{y}_{m}\ ,\ \ \ 
\ddot{y}_{\set{U}}\hspace{-0mm}\triangleq\hspace{-0mm}\sum^M_{m=1} 
(\dot{\Sigma}^{m}_{\set{U}})^{\top} \dot{R}_m \dot{y}_{m}\vspace{0mm} \\
\displaystyle\ddot{\Sigma}_{\set{S}\set{S}}\hspace{-0mm}\triangleq &\hspace{-2mm} \displaystyle{\Sigma}_{\set{S}\set{S}}+\sum^M_{m=1}(\dot{\Sigma}^{m}_{\set{S}})^{\top}\dot{R}_m \dot{\Sigma}^{m}_{\set{S}}\vspace{-0mm}\\
\displaystyle\ddot{\Sigma}_{\set{U}\set{S}}\hspace{-0mm}\triangleq &\hspace{-2mm}\displaystyle\sum^M_{m=1} (\dot{\Sigma}^{m}_{\set{U}})^{\top}\dot{R}_m \dot{\Sigma}^{m}_{\set{S}}\ , \ \ \
\ddot{\Sigma}_{\set{U}\set{U}}\hspace{-0mm}\triangleq\hspace{-0mm}\sum^M_{m=1}(\dot{\Sigma}^{m}_{\set{U}})^{\top} \dot{R}_m \dot{\Sigma}^{m}_{\set{U}} .\vspace{-1.1mm}
\end{array}
$$
\label{def:gs}
\end{defi}
\begin{theorem} 
%
For $B>0$, ${\mu}^{\text{\tiny\emph{LMA}}}_{{\set{U}}}$ \eqref{vapmu} and ${\Sigma}^{\text{\tiny\emph{LMA}}}_{{\set{U}}{\set{U}}}$ \eqref{vapvar}
can be reduced to
${\mu}^{\text{\tiny\emph{LMA}}}_{\set{U}}
=\mu_{\set{U}}+\ddot{y}_{\set{U}} -\ddot{\Sigma}_{\set{U}\set{S}}\ddot{\Sigma}_{\set{S}\set{S}}^{-1}\ddot{y}_{\set{S}}$ and
$
{\Sigma}^{\text{\tiny\emph{LMA}}}_{{\set{U}}{\set{U}}}
={\Sigma}_{\set{U}{\set{U}}} -
\ddot{\Sigma}_{\set{U}\set{U}} +\ddot{\Sigma}_{\set{U}\set{S}}\ddot{\Sigma}_{\set{S}\set{S}}^{-1}\ddot{\Sigma}^{\top}_{\set{U}\set{S}}
$.\vspace{-0mm}
\label{thm2}
\end{theorem}
Its proof in\if\myproof1 Appendix~\ref{pthm2} \fi\if\myproof0 \cite{AA14} \fi essentially relies on the sparsity\vspace{-0mm} of 
$\overline{R}^{-1}_{\set{D}\set{D}}$ and the matrix inversion lemma.\vspace{0.5mm} 

\noindent
\textit{Remark} $1$. To parallelize LMA, each machine/core $m$ constructs and uses the $m$-th local summary to compute the $m$-th summation terms in the global summary, which are then communicated to a master node. The master node constructs and communicates the global summary to the $M$ machines/cores, specifically, by sending the tuple $(\ddot{y}_{\set{S}},\ddot{y}_{\set{U}_m},\ddot{\Sigma}_{\set{S}\set{S}},\ddot{\Sigma}_{\set{U}_m\set{S}},\ddot{\Sigma}_{\set{U}_m\set{U}_m})$ to each machine/core $m$. Finally, each machine/core $m$ uses this received tuple 
to predict the unobserved outputs for the set $\set{U}_m$ of inputs and provide their corresponding predictive uncertainties using ${\mu}^{\text{\tiny{LMA}}}_{{\set{U}_m}}$ \eqref{vapmu} and ${\Sigma}^{\text{\tiny{LMA}}}_{{\set{U}_m}{\set{U}_m}}$ \eqref{vapvar}, respectively.
Computing 
$\overline{\Sigma}_{\set{D}_{m}\set{U}}\vspace{-0mm}$ and $\overline{\Sigma}_{\set{D}^B_{m}\set{U}}$ terms in the local summary
can also be parallelized due to their recursive definition (i.e., \eqref{ares} and \eqref{acov}), as discussed in\if\myproof1 Appendix~\ref{app4}. \fi\if\myproof0 \cite{AA14}. \fi
This parallelization capability of LMA shows another key advantage over existing sparse GP regression methods\footnote{A notable exception is the work of \citeauthor{LowUAI13}~\shortcite{{LowUAI13}} that parallelizes PIC. As mentioned earlier, our LMA generalizes PIC.} in gaining scalability.\vspace{0.5mm}

\noindent
\textit{Remark} $2$. Supposing $M,|\set{U}|,|\set{S}|\leq |\set{D}|$,
LMA can compute ${\mu}^{\text{\tiny{LMA}}}_{{\set{U}}}$ and $\text{tr}({\Sigma}^{\text{\tiny{LMA}}}_{{\set{U}}{\set{U}}})\vspace{-0mm}$ 
distributedly in 
%
%
$\set{O}(|\set{S}|^3+(B|\set{D}|/M)^3+|\set{U}|(|\set{D}|/M)(|\set{S}|+B|\set{D}|/M))$ time
using $M$ parallel machines/cores and sequentially in
$\set{O}(|\set{D}||\set{S}|^2+B|\set{D}|(B|\set{D}|/M)^2+|\set{U}||\set{D}|(|\set{S}|+B|\set{D}|/M))$ time on a single centralized machine.
So, our LMA method incurs cubic time in support set size $|\set{S}|$ and Markov order $B$. 
Increasing the number $M$ of parallel machines/cores and blocks reduces the incurred time of our parallel and centralized LMA methods, respectively.
Without considering communication latency, the speedup\footnote{Speedup is the incurred time of a sequential/centralized algorithm divided by that of its parallel counterpart.} of our parallel LMA method grows with increasing $M$ and training data size $\sset{D}\vspace{-0mm}$; to explain the latter, unlike the additional $\set{O}(\sset{D}\sset{S}^2)$ time of our centralized LMA method that increases with more data, parallel LMA does not have a corresponding $\set{O}((\sset{D}/M)\sset{S}^2)$ term.\vspace{0.5mm} 

\noindent
\textit{Remark} $3$. Predictive performance of LMA is improved by increasing the support set size $|\set{S}|$ and/or Markov order $B$ at the cost of greater time overhead.
From Remark $2$, since LMA incurs cubic time in $|\set{S}|$ as well as in $B$, one should trade off between $|\set{S}|$ and $B$ to reduce the computational cost while achieving the desired predictive performance. In contrast, PIC \cite{LowUAI13,snelson07} (sparse spectrum GP \cite{candela10}) can only vary support set size (number of spectral points) to obtain the desired predictive performance.\vspace{0.5mm}

\noindent
\textit{Remark} $4$. 
We have illustrated through a simple toy example in\if\myproof1 Appendix~\ref{1dtoy} \fi\if\myproof0 \cite{AA14} \fi that, unlike the   local GPs approach, 
LMA does not exhibit any discontinuity in its predictions despite data partitioning.
%
\section{Experiments and Discussion}
\label{expts}
This section first empirically evaluates the predictive performance and scalability of our proposed centralized and parallel LMA methods against that of the state-of-the-art centralized PIC
\cite{snelson07}, parallel PIC \cite{LowUAI13}, \emph{sparse spectrum GP} (SSGP) \cite{candela10}, and FGP 
on two real-world datasets:
(a) The SARCOS dataset \cite{Vijayakumar05} of size $48933$ is obtained from an inverse dynamics problem for a $7$ degrees-of-freedom SARCOS robot
arm. Each input is specified by a $21$D feature vector of joint positions, velocities, and accelerations. The output corresponds to one of the $7$ joint torques. 
(b) The AIMPEAK dataset \cite{LowUAI13} of size $41850$ comprises traffic speeds (km/h) along $775$ road segments of an urban road network during morning peak hours 
on April $20$, $2011$. 
Each input (i.e., road segment) denotes a $5$D feature vector of
length, number of lanes, speed limit, direction, and time. The time
dimension comprises $54$ five-minute time slots. This traffic dataset is modeled using a relational GP \cite{chen12} whose correlation structure can exploit the road segment features and road network topology information.
The outputs correspond to the traffic speeds.

Both datasets are modeled using GPs whose prior covariance $\sigma_{xx'}\vspace{-0mm}$ is defined by
the squared exponential covariance function\footnote{For AIMPEAK dataset, multi-dimensional scaling is used to map the input domain (i.e., of road segments) onto the Euclidean space \cite{chen12} before applying the covariance function.}
%
$\vspace{-0mm}\sigma_{xx'}\triangleq \sigma_s^2\exp({-0.5\sum_{i=1}^{d}
    (x_i-x'_i)^2/\ell_i^2})+\sigma_n^2\delta_{xx'}$
%
where $x_i \left(x'_i \right)$ is the $i$-th component of input
feature vector $x \left(x'\right)$, the hyperparameters
$\sigma^2_s,\sigma^2_n,\ell_1,\dots,\ell_d$ are, respectively, signal
variance, noise variance, and length-scales, and $\delta_{x x'}$ is a
Kronecker delta that is $1$ if $x = x'$ and $0$ otherwise.  
The hyperparameters are learned using randomly selected data of size $10000$ via maximum likelihood estimation.
Test data of size $\sset{U}=3000$ are randomly selected from each dataset for predictions.
From remaining data, training data
of varying $\sset{D}$
are randomly selected. 
%
Support sets for LMA and PIC and the set  $\mathcal{S}$ of spectral points for SSGP are selected randomly from both datasets\footnote{Varying the set $\mathcal{S}$ of spectral points over $50$ random instances hardly changes the predictive performance of SSGP in our experiments because a very large set of spectral points ($|\mathcal{S}|=4096$) is used in order to achieve predictive performance as close as possible to FGP and our LMA method (see Table~\ref{tab:sarcos}).}.

The experimental platform is a cluster of $32$ computing nodes connected
via gigabit links: Each node runs a Linux system with Intel$\circledR$ Xeon$\circledR$ E$5620$ at $2.4$~GHz with $24$~GB memory and $16$ cores.
Our parallel LMA method and parallel PIC 
are tested with different numbers $M= 32$, $48$, and $64$ of cores;
all $32$ computing nodes with $1$, $1$-$2$, and $2$ cores each are used, respectively.
For parallel LMA and parallel PIC, each computing node will be storing, respectively, a subset of the training data $(\set{D}_m\cup\set{D}^B_m, y_{\set{D}_m\cup\set{D}^B_m})\vspace{-0mm}$ and $(\set{D}_m, y_{\set{D}_m})$ associated with its own core $m$. 
%
%
%
%

Three performance metrics are used to evaluate the tested methods: 
(a) Root mean square error (RMSE)\vspace{-0mm} $(\sset{U}^{-1}\sum_{x\in
\set{U}}(y_x-\mu_{x|\set{D}})^2)^{1/2}$, 
(b) incurred time, and (c) speedup. For RMSE metric, each tested method has to plug its predictive mean into $\mu_{x|\set{D}}$. 
\begin{table}
{\tiny
\begin{tabular}{c}
\begin{tabular}{|l|cccc|}
\hline
$|\set{D}|$ & $8000$ & $16000$ & $24000$ & $32000$\\
\hline
FGP & $2.4(285)$ & $2.2(1799)$ & $2.1(5324)$ & $2.0(16209)$ \\
SSGP & $2.4(2029)$ & $2.2(3783)$ & $2.1(5575)$ & $2.0(7310)$ \\
\hline
\multicolumn{5}{|c|}{$M=32$}\\
\hline
{\bf LMA} & $2.4(56)$ & $2.2(87)$ & $2.1(157)$ & $2.0(251)$ \\
PIC & $2.4(254)$ & $2.2(294)$ & $2.1(323)$ & $2.0(363)$ \\
\hline
\multicolumn{5}{|c|}{$M=48$}\\
\hline
{\bf LMA} & $2.4(51)$ & $2.2(84)$ & $2.1(126)$ & $2.0(192)$ \\
PIC & $2.4(273)$ & $2.2(308)$ & $2.1(309)$ & $2.0(332)$ \\
\hline
\multicolumn{5}{|c|}{$M=64$}\\
\hline
{\bf LMA} & $2.4(61)$ & $2.2(87)$ & $2.1(111)$ & $2.0(155)$ \\
PIC & $2.4(281)$ & $2.2(286)$ & $2.1(290)$ & $2.0(324)$ \\
\hline
\end{tabular}
\vspace{0.5mm}\\
(a) Parallel LMA ($B=1$, $|\set{S}|=2048$), parallel PIC ($|\set{S}|=4096$), SSGP ($|\set{S}|=4096$)\vspace{1mm}\\
\begin{tabular}{|l|cccc|}
\hline
$|\set{D}|$ & $8000$ & $16000$ & $24000$ & $32000$\\
\hline
FGP & $7.9(271)$ & $7.3(1575)$ & $7.0(5233)$ & $6.9(14656)$ \\
SSGP & $8.1(2029)$ & $7.5(3781)$ & $7.3(5552)$ & $7.2(7309)$ \\
\hline
\multicolumn{5}{|c|}{$M=32$}\\
\hline
{\bf LMA} & $8.4(20)$ & $7.5(44)$ & $7.1(112)$ & $6.9(216)$ \\
PIC & $8.1(484)$ & $7.5(536)$ & $7.3(600)$ & $7.2(598)$ \\
\hline
\multicolumn{5}{|c|}{$M=48$}\\
\hline
{\bf LMA} & $8.4(18)$ & $7.5(33)$ & $7.0(74)$ & $6.8(120)$ \\
PIC & $8.1(542)$ & $7.5(590)$ & $7.3(598)$ & $7.2(616)$ \\
\hline
\multicolumn{5}{|c|}{$M=64$}\\
\hline
{\bf LMA} & $8.4(17)$ & $7.5(28)$ & $7.0(57)$ & $6.7(87)$ \\
PIC & $8.1(544)$ & $7.5(570)$ & $7.3(589)$ & $7.2(615)$ \\
\hline
\end{tabular}\vspace{0.5mm}\\
(b) Parallel LMA ($B=1$, $|\set{S}|=1024$), parallel PIC ($|\set{S}|=5120$), SSGP ($|\set{S}|=4096$)
\end{tabular}
\vspace{-2mm}}
\caption{RMSEs and incurred times (seconds) reported in brackets of parallel LMA, parallel PIC, SSGP, and FGP with varying data sizes $|\set{D}|$ and numbers $M$ of cores for (a) SARCOS and (b) AIMPEAK datasets.
\vspace{-4.4mm}}
\label{tab:sarcos}
\label{tab:aimpeak}
\end{table}
%

Table~\ref{tab:sarcos}
shows results of RMSEs and incurred times  of parallel LMA, parallel PIC, SSGP, and FGP averaged over $5$ random instances with varying data sizes $|\set{D}|$ and cores $M$ for both datasets.
The observations are as follows:

\noindent
(a) Predictive performances of all tested methods improve with more data, which is expected.
For SARCOS dataset, parallel LMA, parallel PIC, and SSGP achieve predictive performances comparable to that of FGP. For AIMPEAK dataset, parallel LMA does likewise and outperforms parallel PIC and SSGP with more data ($|\set{D}|\geq 24000$), which may be due to its more relaxed conditional independence assumption with larger data (Remark $1$ after Proposition~\ref{prop1}).

\noindent
(b) The incurred times of all tested methods increase with more data, which is also expected. FGP scales very poorly with larger data such that it incurs $>4$ hours for $|\set{D}|= 32000$. In contrast, parallel LMA incurs only $1$-$5$ minutes for both datasets when $|\set{D}|= 32000$. Parallel LMA incurs much less time than parallel PIC 
and SSGP
while achieving a comparable or better predictive performance
because it requires a significantly smaller $|\set{S}|$ than parallel PIC and SSGP simply by imposing a $1$-order Markov property ($B=1$) on the residual process (Remark $3$ after Theorem~\ref{thm2}). 
Though $B$ is only set to $1$,
the dense residual covariance matrix approximation provided by LMA (as opposed to sparse approximation of PIC) is good enough to achieve its predictive performances reported in Table~\ref{tab:sarcos}. 
From Table~\ref{tab:aimpeak}b, when training data is small ($|\set{D}|= 8000$) for AIMPEAK dataset, parallel PIC incurs more time than FGP due to its huge $|\set{S}|=5120$, which causes communication latency to dominate the incurred time \cite{LowUAI13}. When $|\set{D}|\leq 24000$, SSGP also incurs more time than FGP due to its large $|\set{S}|=4096$.

\noindent
(c) Predictive performances of parallel LMA and PIC generally remain stable with more cores, thus justifying the practicality of their structural assumptions to gain time efficiency.

Table~\ref{tab:aimpeaks} shows results of speedups of parallel LMA and parallel PIC as well as incurred times of their \emph{centralized} counterparts averaged over $5$ random instances with varying data sizes $|\set{D}|$ and numbers $M$ of cores for AIMPEAK dataset. The observations are as follows:

\noindent
(a) The incurred times of centralized LMA and centralized PIC increase with more data, which is expected.
When $|\set{D}|\geq 32000$, centralized LMA incurs only $16$-$30$ minutes
(as compared to FGP incurring $>4$ hours) while centralized PIC and SSGP incur, respectively, more than $3.5$ and $2$ hours due to their huge $|\set{S}|$. In fact,  Table~\ref{tab:aimpeaks} shows that centralized PIC incurs even more time than FGP for almost all possible settings of $|\set{D}|$ and $M$ due to its huge support set.

\noindent
(b) The speedups of parallel LMA and parallel PIC generally increase with more data, as explained in Remark $2$ after Theorem~\ref{thm2}, except for that of parallel LMA being slightly higher than expected when $|\mathcal{D}|=16000$.

\noindent
(c) The incurred time of centralized LMA decreases with more blocks (i.e., larger $M$), as explained in Remark $2$ after Theorem~\ref{thm2}.
This is also expected of centralized PIC, but its incurred time increases with more blocks instead due to its huge support set, which entails large-scale matrix operations causing a huge number of cache misses\footnote{A cache miss causes the processor to access the data from main memory, which costs $10\times$ more time than a cache memory access.}. This highlights the need to use a sufficiently small support set on a single centralized machine so that cache misses will contribute less to incurred time, as compared to data processing.

\noindent
(d) The speedup of parallel LMA increases with more cores, as explained in Remark $2$ after Theorem~\ref{thm2}. 
Though the speedup of parallel PIC appears to increase considerably with more cores, it is primarily due to the substantial number of cache misses (see observation c above) that inflates the incurred time of centralized PIC excessively. 
%
%
\begin{table}
{\tiny
\begin{tabular}{|l|cccc|}
\hline
$|\set{D}|$ & $8000$ & $16000$ & $24000$ & $32000$\\
\hline
FGP & $-(271)$ & $-(1575)$ & $-(5233)$ & $-(14656)$ \\
SSGP & $-(2029)$ & $-(3781)$ & $-(5552)$ & $-(7309)$ \\
\hline
\multicolumn{5}{|c|}{$M=32$}\\
\hline
{\bf LMA} & $6.9(139)$ & $9.4(414)$ & $8.0(894)$ & $8.2(1764)$ \\
PIC &  $19.4(9432)$ & $18.8(10105)$ & $19.3(11581)$ & $21.6(12954)$ \\
\hline
\multicolumn{5}{|c|}{$M=48$}\\
\hline
{\bf LMA} & $6.9(125)$ & $10.2(338)$ & $9.2(678)$ & $10.2(1227)$ \\
PIC & $25.3(13713)$ & $24.1(14241)$ & $26.2(15684)$ & $26.8(16515)$ \\
\hline
\multicolumn{5}{|c|}{$M=64$}\\
\hline
{\bf LMA} & $7.1(120)$ & $10.8(302)$ & $10.1(576)$ & $11.5(1003)$ \\
PIC & $31.6(17219)$ & $31.5(17983)$ & $33.0(19469)$ & $33.3(20503)$ \\
\hline
\end{tabular}\vspace{-3mm}
\caption{Speedups of 
parallel LMA ($B=1$, $|\set{S}|=1024$) and parallel PIC ($|\set{S}|=5120$)
and incurred times (seconds) reported in brackets of their \emph{centralized} counterparts with varying data sizes $|\set{D}|$ and cores $M$ for AIMPEAK dataset.\vspace{-3mm}}
\label{tab:aimpeaks}
}
\end{table}
\begin{figure}
\begin{tabular}{cc}
\hspace{-3mm}
\includegraphics[height=32mm]{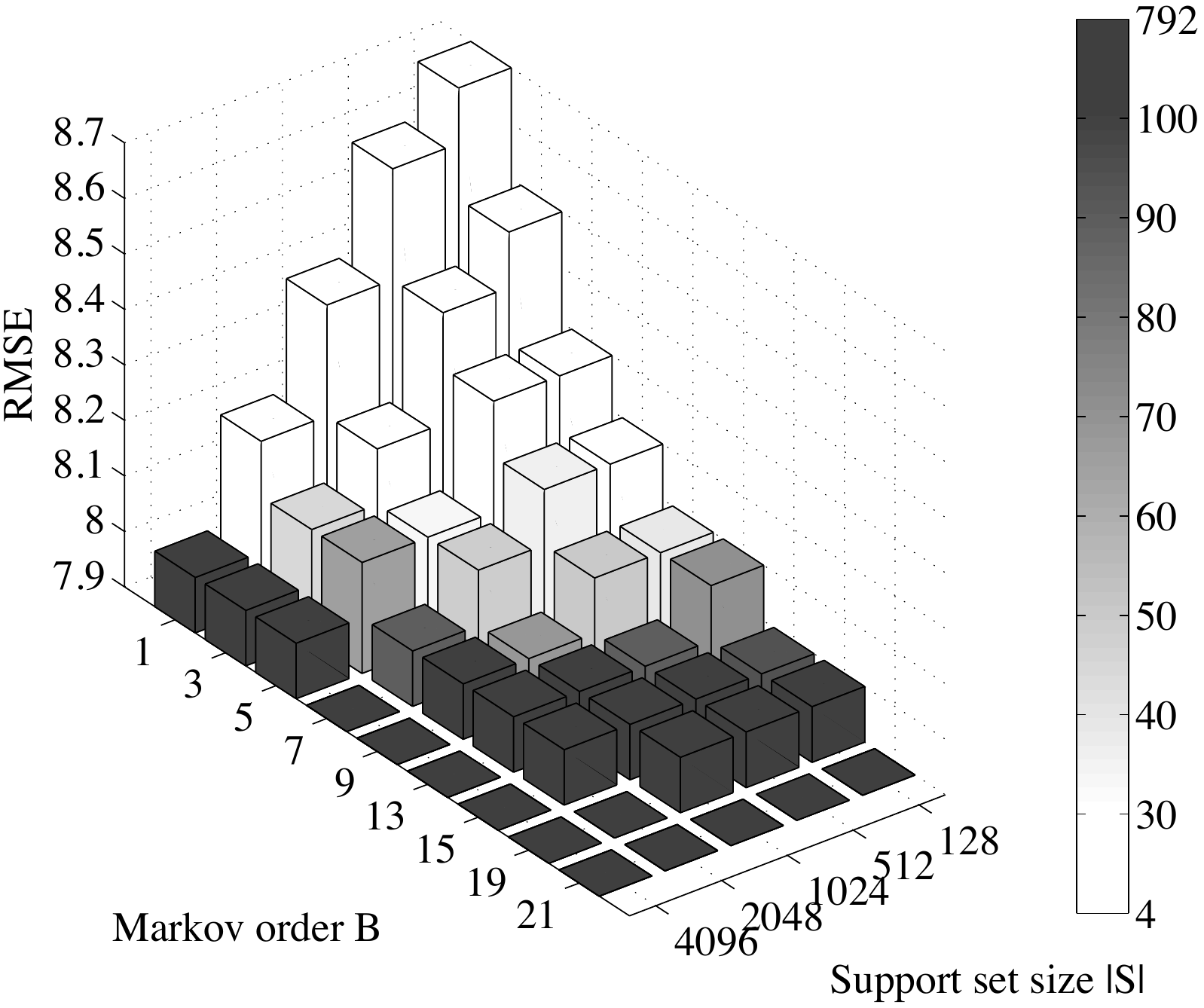} & \hspace{-4.5mm}\includegraphics[height=32mm]{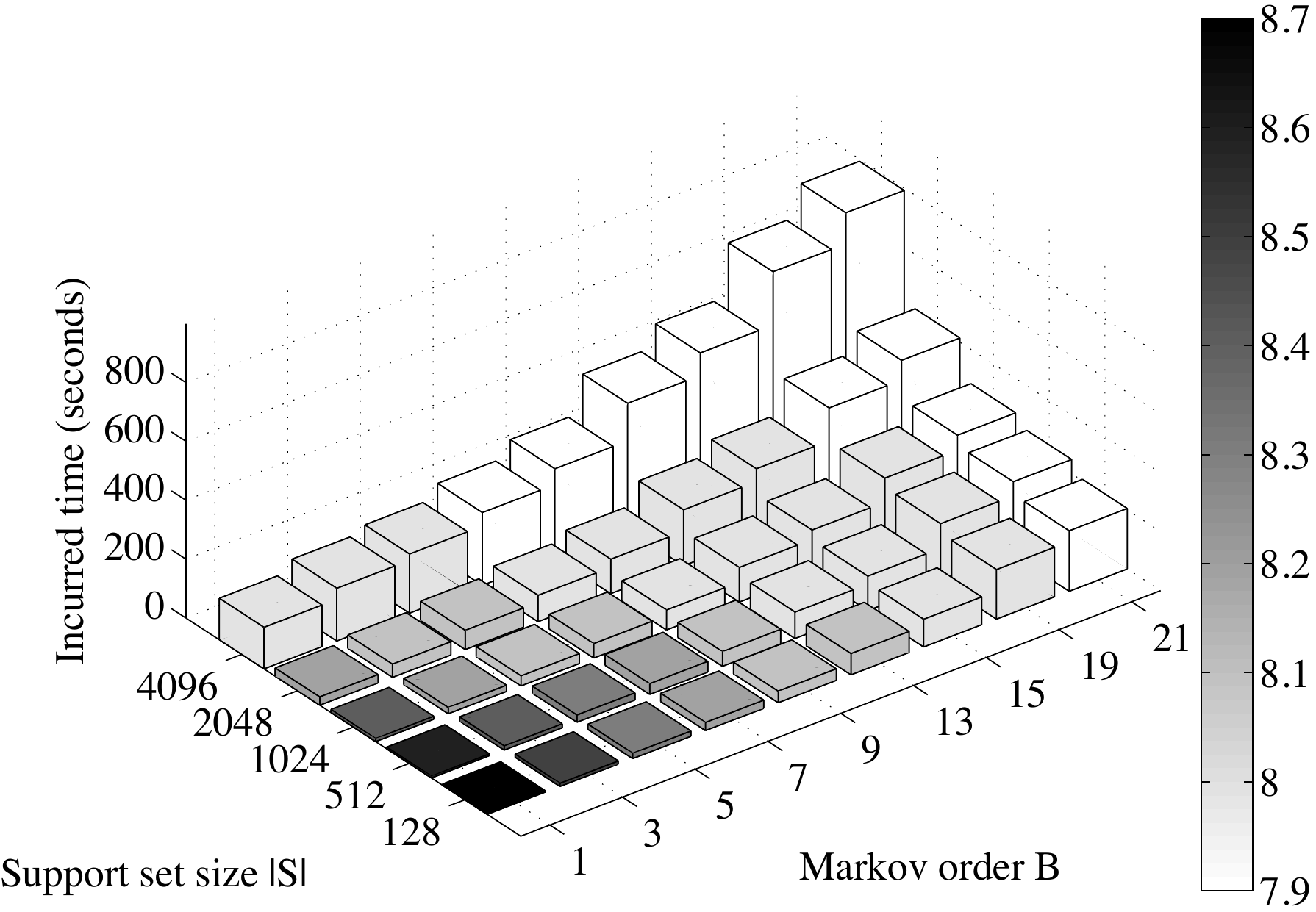}\vspace{-4mm}
\end{tabular}
\caption{RMSEs and incurred times (seconds) of 
parallel LMA with varying support set sizes $|\mathcal{S}|=128,512,1024,2048,4096$ and Markov orders $B=1,3,5,7,9,13,15,19,21$, $|\mathcal{D}|=8000$, and $M=32$ for AIMPEAK dataset. Darker gray implies longer incurred time (larger RMSE) for the left (right) plot.\vspace{-4.6mm}}
\label{varysb}
\end{figure}

Fig.~\ref{varysb} shows results of RMSEs and incurred times of parallel LMA averaged over $5$ random instances with varying support set sizes $|\mathcal{S}|$ and Markov orders $B$, $|\mathcal{D}|=8000$, and $M=32$ obtained using $8$ computing nodes (each using $4$ cores) for AIMPEAK dataset. Observations are as follows:

\noindent
(a) To achieve RMSEs of $8.1$ and $8.0$ with least incurred times, one should trade off a larger support set size $|\mathcal{S}|$ for a larger Markov order $B$ (or vice versa) to arrive at the respective settings of $|\mathcal{S}|=1024, B=5$ ($34$ seconds) and $|\mathcal{S}|=1024, B=9$ ($68$ seconds), which agrees with Remark $3$ after Theorem~\ref{thm2}. However, to achieve the same RMSE of $7.9$ as FGP, the setting of $|\mathcal{S}|=128, B=21$ incurs the least time (i.e., $205$ seconds), which seems to indicate that, with small data ($|\mathcal{D}|=8000$), we should instead focus on increasing Markov order $B$ for LMA to achieve the same predictive performance as FGP; recall that when $B=M-1$, LMA becomes FGP. This provides an empirically cheaper and more reliable alternative to increasing $|\mathcal{S}|$ for achieving predictive performance comparable to FGP, the latter of which, in our experiments, causes Cholesky factorization failure easily when $|\mathcal{S}|$ becomes excessively large.

\noindent
(b) When $|\mathcal{S}|=1024$, $B=1$, and $M=32$, parallel LMA using $8$ computing nodes incurs less time (i.e., $10$ seconds) than that using $32$ nodes (i.e., $20$ seconds; see Table~\ref{tab:sarcos}b) because the communication latency between cores within a machine is significantly less than that between machines.

Next, the predictive performance and scalability of our parallel LMA method are empirically compared with that of parallel PIC using the large \emph{EMULATE mean sea level pressure} (EMSLP) dataset \cite{Ansell06} 
of size $1278250$ on a $5^\circ$ lat.-lon. grid bounded within lat. $25$-$70$N and lon. $70$W-$50$E from $1900$ to $2003$. Each input denotes a $6$D feature vector of latitude, longitude, year, month, day, and incremental day count (starting from $0$ on first day).
The output is the mean sea level pressure (Pa). 
The experimental setup is the same as before, except for 
the platform that is a cluster of $16$ computing nodes connected via gigabit links: Each node runs a Linux system with AMD Opteron$^{\text{TM}}$ $6272$ at $2.1$ GHz with $32$ GB memory and $32$ cores. 
\begin{table}
{\tiny
\begin{tabular}{|l|ccccc|}
\hline
$|\set{D}|$ & $128000$ & $256000$ & $384000$ & $512000$ & $1000000$\\
\hline
{\bf LMA} & $823(155)$ & $774(614)$ & $728(3125)$ & $682(7154)$ & $506(78984)$ \\
PIC & $836(948)$ & $-(-)$ & $-(-)$ & $-(-)$ & $-(-)$ \\
\hline
\end{tabular}\vspace{-2.9mm}}
\caption{RMSEs and incurred times (seconds) reported in brackets of parallel LMA ($B=1$, $|\set{S}|= 512$) and parallel PIC ($|\set{S}|=3400$) with $M=512$ cores and varying data sizes $|\set{D}|$ for EMSLP dataset.\vspace{-5mm}
}
\label{tab:apart}
\end{table}

Table~\ref{tab:apart} shows results of RMSEs and incurred times of parallel LMA and parallel PIC averaged over $5$ random instances with $M=512$ cores and varying data sizes $|\set{D}|$ for EMSLP dataset. When $|\set{D}|=128000$, parallel LMA incurs much less time than parallel PIC while achieving better predictive performance because it requires a significantly smaller $|\set{S}|$ by setting $B=1$, as explained earlier. When $|\set{D}|\geq 256000$, parallel PIC fails due to insufficient shared memory between cores. On the other hand, parallel LMA does not experience this issue and incurs from $10$ minutes for $|\set{D}|=256000$ to about $22$ hours for $|\set{D}|=1000000$.\vspace{0.5mm}


\noindent
{\bf Summary of Experimental Results.}
LMA is significantly more scalable than FGP in the data size while achieving a comparable predictive performance for SARCOS and AIMPEAK datasets.
For example, when $|\set{D}|= 32000$ and $M\geq 48$, our centralized and parallel LMA methods are, respectively, at least $1$ and $2$ orders of magnitude faster than FGP while achieving comparable predictive performances for AIMPEAK dataset.
Our centralized (parallel) LMA method also incurs much less time than centralized PIC (parallel PIC) and SSGP while achieving comparable or better predictive performance because LMA requires a considerably smaller support set size $|\set{S}|$ than PIC and SSGP simply by setting Markov order $B=1$, as explained earlier.
Trading off between support set size and Markov order of LMA results in less incurred time while achieving the desired predictive performance.
LMA gives a more reliable alternative of increasing the Markov order (i.e., to increasing support set size) for achieving predictive performance similar to FGP; in practice, a huge support set causes Cholesky factorization failure and insufficient shared memory between cores easily. Finally, parallel LMA can scale up to work for EMSLP dataset of more than a million in size.
\vspace{-2.1mm}
\section{Conclusion}
\vspace{-0.5mm}
This paper describes a LMA method that leverages the dual computational advantages stemming from complementing the low-rank covariance
matrix approximation based on support set with the dense residual covariance matrix approximation due to Markov assumption.
As a result, LMA can make a more relaxed conditional independence assumption (especially with larger data) than many existing sparse GP regression methods utilizing low-rank representations, the latter of which utilize a sparse residual covariance matrix approximation.
Empirical results have shown that our centralized (parallel) LMA method is much more scalable than FGP and time-efficient than centralized PIC (parallel PIC) and SSGP while achieving comparable predictive performance.
In our future work, we plan to develop a technique to automatically determine the ``optimal'' support set size and Markov order and devise an ``anytime'' variant of LMA using stochastic variational inference like \cite{Lawrence13} so that it can train with a small subset of data in each iteration instead of learning using all the data. We also plan to release the source code at http://code.google.com/p/pgpr/.\vspace{1mm}

\noindent
{\bf Acknowledgments.}
This work was supported by Singapore-MIT Alliance for Research and Technology Subaward Agreement No. $52$ R-$252$-$000$-$550$-$592$.\vspace{-1mm}

\bibliographystyle{aaai}
\bibliography{traffic}

\if \myproof1
\clearpage
\appendix
\section{Proof of Theorem~\ref{thm1}}
\label{pthm1}
$$
\begin{array}{l}
D_\text{KL}({R}_{\set{D}\set{D}},\overline{R}_{\set{D}\set{D}})+D_\text{KL}(\overline{R}_{\set{D}\set{D}},\widehat{R})\\
\displaystyle= \frac{1}{2}\hspace{-0.5mm}\left(\text{tr}({R}_{\set{D}\set{D}}\overline{R}_{\set{D}\set{D}}^{-1})\hspace{-0.5mm}-\hspace{-0.5mm}\log|{R}_{\set{D}\set{D}}\overline{R}_{\set{D}\set{D}}^{-1}|\hspace{-0.5mm}-\hspace{-0.5mm}|\set{D}|\right)\hspace{-0.5mm}+\vspace{0.5mm}\\
\quad
\displaystyle\frac{1}{2}\hspace{-0.5mm}\left(\text{tr}(\overline{R}_{\set{D}\set{D}}\widehat{R}^{-1})\hspace{-0.5mm}-\hspace{-0.5mm}\log|\overline{R}_{\set{D}\set{D}}\widehat{R}^{-1}|\hspace{-0.5mm}-\hspace{-0.5mm}|\set{D}|\right)\vspace{1mm}\\
\displaystyle= \frac{1}{2}\left( \text{tr}(\overline{R}_{\set{D}\set{D}}\widehat{R}^{-1}) -  \log|{R}_{\set{D}\set{D}}| - \log|\widehat{R}^{-1}| -|\set{D}| \right)\vspace{1mm}\\
= \displaystyle\frac{1}{2}\left(\text{tr}({R}_{\set{D}\set{D}}\widehat{R}^{-1})-\log|{R}_{\set{D}\set{D}}\widehat{R}^{-1}|-|\set{D}|\right)\\
= D_\text{KL}({R}_{\set{D}\set{D}},\widehat{R})\ .
\end{array}
$$
The second equality is due to $\text{tr}({R}_{\set{D}\set{D}}\overline{R}^{-1}_{\set{D}\set{D}})=\text{tr}(\overline{R}_{\set{D}\set{D}}\overline{R}^{-1}_{\set{D}\set{D}})=\text{tr}(I_{|\set{D}|})=|\set{D}|$, which follows from the observations that the blocks within the $B$-block bands of ${R}_{\set{D}\set{D}}$ and $\overline{R}_{\set{D}\set{D}}$ are the same and $\overline{R}^{-1}_{\set{D}\set{D}}$ is $B$-block-banded (Proposition~\ref{prop1}). The third equality follows from the first observation above and the definition that $\widehat{R}^{-1}$ is $B$-block-banded. 
Since $D_\text{KL}(\overline{R}_{\set{D}\set{D}},\widehat{R})\geq 0$, $D_\text{KL}({R}_{\set{D}\set{D}},\widehat{R})\geq D_\text{KL}({R}_{\set{D}\set{D}},\overline{R}_{\set{D}\set{D}})$.
\section{Proof of Theorem~\ref{thm2}}
\label{pthm2}
The following lemma is necessary for deriving our main result here. It shows that the sparsity of $B$-block-banded $\overline{R}^{-1}_{\set{D}\set{D}}$ (Proposition~\ref{prop1}) extends to that of its Cholesky factor (Fig.~\ref{fig:matrix2}):
\begin{lemma}
Let $\overline{R}^{-1}_{\set{D}\set{D}} \triangleq U^{\top} U$ where Cholesky factor $U=[U_{m{n}}]_{m,n=1,\ldots, M}$ is an upper triangular matrix (Fig.~\ref{fig:matrix2}).
Then,
$U_{m{n}}=\underline{0}$ 
if $m-n>0$ or $n-m>B$.
Furthermore, for $m=1,\ldots, M$,
$U_{mm} = \text{\em cholesky}(\dot{R}_m)$ and $U^B_{m}\triangleq [U_{mn}]_{n=m+1,\ldots,\min(m+B,M)} = - U_{mm} R_{\set{D}_m\set{D}^B_{m}}R_{\set{D}^B_{m}\set{D}^B_{m}}^{-1}$.
%
%
\label{lemma1}
\end{lemma}
Its proof follows directly from block-banded matrix results of \citeauthor{Moura05}~\shortcite{Moura05}\vspace{-0.5mm} (i.e., Lemma~$1.1$ and Theorem $1$).
\begin{figure}[b]
\includegraphics[scale=0.2]{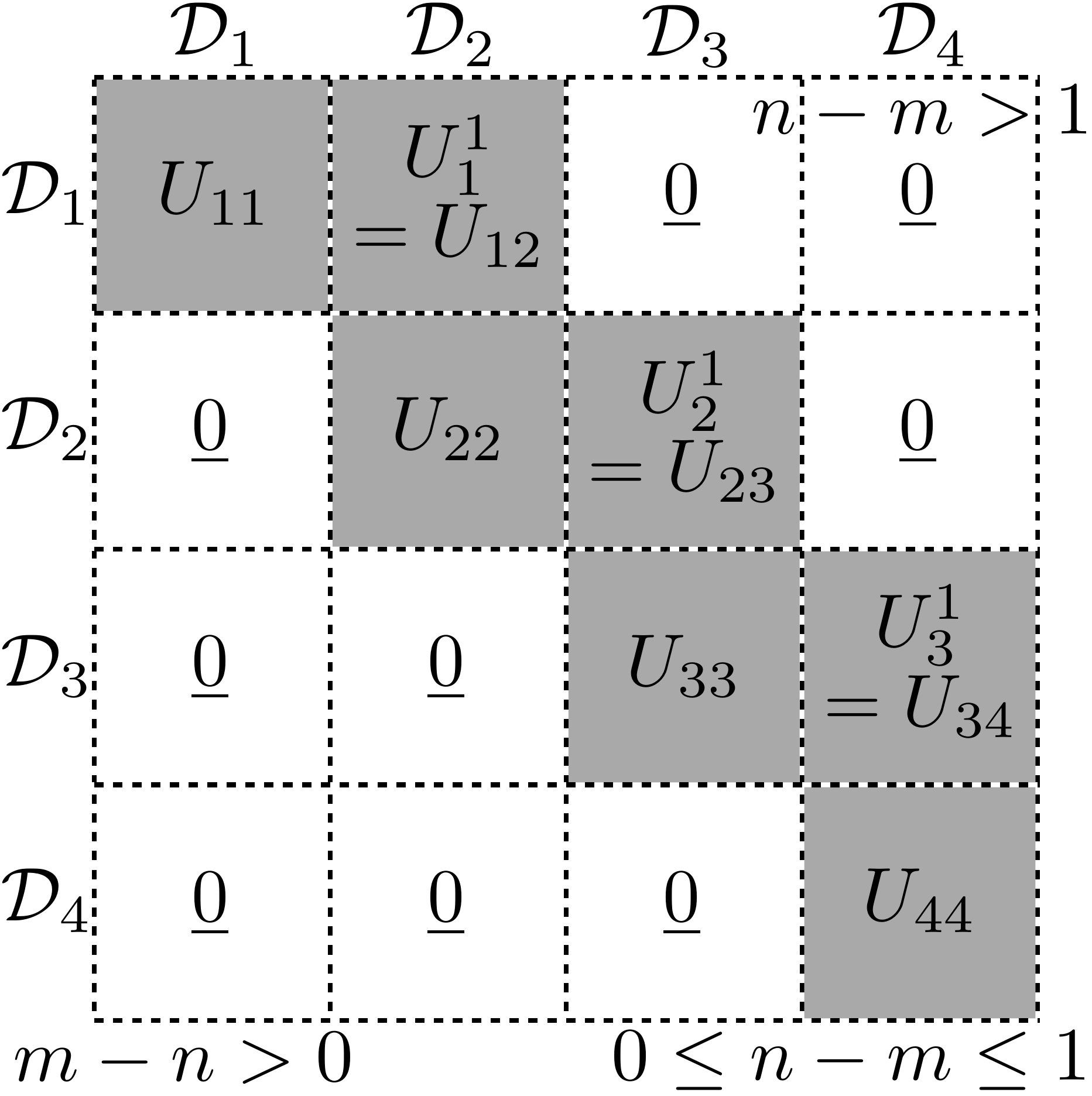}\vspace{-1mm}\\
\caption{Cholesky factor $U$ with $B=1$ and $M=4$. Unshaded blocks outside $B$-block band of $\overline{R}^{-1}_{\set{D}\set{D}}$ (i.e., $|m-n|> B$) are $\underline{0}$ (Fig.~\ref{fig:matrix}b), which result in the unshaded blocks of its Cholesky factor $U$ being $\underline{0}$ (i.e., $m-n>0$ or $n-m>B$).}\vspace{-0mm}
\label{fig:matrix2}
\end{figure}
%
\begin{equation} 
\begin{array}{l}
  \overline{\Sigma}_{\set{D}\set{D}}^{-1}\\ 
  \displaystyle =   \left(\Sigma_{\set{D}\set{S}}\Sigma^{-1}_{\set{S}\set{S}}\Sigma_{\set{S}\set{D}} + \overline{R}_{\set{D}\set{D}}\right)^{-1}\\
  \displaystyle = \overline{R}^{-1}_{\set{D}\set{D}}\hspace{-0.8mm} -\hspace{-0.8mm}\overline{R}^{-1}_{\set{D}\set{D}}\Sigma_{\set{D}\set{S}}\hspace{-1mm}
  \left(\Sigma_{\set{S}\set{S}}\hspace{-0.8mm}+\hspace{-0.8mm}\Sigma_{\set{S}\set{D}}\overline{R}^{-1}_{\set{D}\set{D}}\Sigma_{\set{D}\set{S}}\right)^{\hspace{-1mm}-1}\hspace{-1.5mm}
  \Sigma_{\set{S}\set{D}}\overline{R}^{-1}_{\set{D}\set{D}}\\ 
\displaystyle
=\overline{R}^{-1}_{\set{D}\set{D}}-\overline{R}^{-1}_{\set{D}\set{D}}\Sigma_{\set{D}\set{S}}\ddot{\Sigma}_{\set{S}\set{S}}^{-1}\Sigma_{\set{S}\set{D}}\overline{R}^{-1}_{\set{D}\set{D}}
\ .  
\end{array} 
\label{eq:pitck} 
\end{equation} 
The second equality is due to the matrix inversion lemma.  The last equality follows from
$$
\hspace{-1.8mm}
\begin{array}{l}
\displaystyle\Sigma_{\set{S}\set{D}} \overline{R}^{-1}_{\set{D}\set{D}}\Sigma_{\set{D}\set{S}}\vspace{1mm}\\
\displaystyle =\Sigma_{\set{S}\set{D}} U^{\top} U \Sigma_{\set{D}\set{S}}\\
\displaystyle =\sum^M_{m=1} \Sigma_{\set{S}(\set{D}_{m}\cup\set{D}^B_{m})}
\left[\hspace{-1mm}
\begin{array}{c}
U^{\top}_{mm}\\
U^{B\top}_{m}
\end{array}
\hspace{-1mm}\right]
\left[U_{mm}, U^B_{m} \right]  \Sigma_{(\set{D}_{m}\cup\set{D}^B_{m})\set{S}}\nonumber\\
= \displaystyle\sum^M_{m=1} \hspace{-1mm}
\left(U_{mm}\Sigma_{\set{D}_{m}\set{S}}+
U^B_{m}\Sigma_{\set{D}^B_{m}\set{S}}\right)^{\hspace{-1mm}\top}\hspace{-1mm}
\left(U_{mm}\Sigma_{\set{D}_{m}\set{S}}+
U^B_{m}\Sigma_{\set{D}^B_{m}\set{S}}\right)\\
= \displaystyle\sum^M_{m=1} (\dot{\Sigma}^{m}_{\set{S}})^{\top}\dot{R}_m  \dot{\Sigma}^{m}_{\set{S}}\\
= \ddot{\Sigma}_{\set{S}\set{S}} - {\Sigma}_{\set{S}\set{S}}\ .
\end{array}
$$
The fourth equality is due to Lemma~\ref{lemma1} and Definition~\ref{def:ls}.
The last equality follows from Definition~\ref{def:gs}.
$$
\begin{array}{l} 
{\mu}^{\text{\tiny {LMA}}}_{\set{U}}\\
\stackrel{\eqref{vapmu}}{=}\displaystyle\mu_{\set{U}}+\overline{\Sigma}_{{\set{U}}\set{D}}\overline{\Sigma}_{\set{D}\set{D}}^{-1}\left(y_\set{D}-\mu_\set{D} \right)\\
\stackrel{\eqref{eq:pitck}}{=}\displaystyle\mu_{\set{U}}+\overline{\Sigma}_{{\set{U}}\set{D}}
\overline{R}^{-1}_{\set{D}\set{D}}\left(y_\set{D}-\mu_\set{D} \right) - \\
\quad\displaystyle\overline{\Sigma}_{{\set{U}}\set{D}}
\overline{R}^{-1}_{\set{D}\set{D}}\Sigma_{\set{D}\set{S}}\ddot{\Sigma}_{\set{S}\set{S}}^{-1}\Sigma_{\set{S}\set{D}}\overline{R}^{-1}_{\set{D}\set{D}}
\left(y_\set{D}-\mu_\set{D} \right)\vspace{0.5mm}\\
=\displaystyle\mu_{\set{U}}+\ddot{y}_{\set{U}} -\ddot{\Sigma}_{\set{U}\set{S}}\ddot{\Sigma}_{\set{S}\set{S}}^{-1}\ddot{y}_{\set{S}}\ .
\end{array}
$$
The last equality is derived using Definition~\ref{def:gs}, specifically, from the expressions of the following three $\ddot{y}_{\set{U}}$, $\ddot{\Sigma}_{\set{U}\set{S}}$, and $\ddot{y}_{\set{S}}$ components, respectively:
$$
\hspace{-1.8mm}
\begin{array}{l}
\displaystyle\overline{\Sigma}_{{\set{U}}\set{D}}
\overline{R}^{-1}_{\set{D}\set{D}}\left(y_\set{D}-\mu_\set{D} \right)\vspace{0.5mm}\\
\displaystyle =\overline{\Sigma}_{{\set{U}}\set{D}} U^{\top} U \left(y_\set{D}-\mu_\set{D} \right)\\
\displaystyle =\sum^M_{m=1} \overline{\Sigma}_{{\set{U}}(\set{D}_{m}\cup\set{D}^B_{m})}\hspace{-1mm}
\left[\hspace{-1mm}
\begin{array}{c}
U^{\top}_{mm}\\
U^{B\top}_{m}
\end{array}
\hspace{-1mm}\right]\left[U_{mm}, U^B_{m} \right]\left(y_{\set{D}_{m}\cup\set{D}^B_{m}}-\mu_{\set{D}_{m}\cup\set{D}^B_{m}} \right)\\
= \displaystyle\sum^M_{m=1} 
\left(U_{mm}\overline{\Sigma}_{{\set{D}_{m}\set{U}}}  +
U^{B}_{m}\overline{\Sigma}_{\set{D}^B_{m}{\set{U}}} \right)^{\hspace{-1mm}\top}\vspace{0.5mm}\\
\quad\quad\quad\left(U_{mm}\left(y_{\set{D}_m}-\mu_{\set{D}_m} \right)+
U^B_{m} \left(y_{\set{D}^B_{m}}-\mu_{\set{D}^B_{m}} \right)\right)\\
=\displaystyle\sum^M_{m=1} 
(\dot{\Sigma}^{m}_{\set{U}})^{\top} \dot{R}_m  \dot{y}_{m}\\
= \ddot{y}_{\set{U}}\ .
\end{array}
$$
The fourth equality is due to Lemma~\ref{lemma1} and Definition~\ref{def:ls}.
$\overline{\Sigma}_{\set{D}_{m}\set{U}}= \left[\overline{\Sigma}_{\set{D}_{m}\set{U}_{n}}\right]_{n=1,\ldots,M}$ and $\overline{\Sigma}_{\set{D}^B_{m}\set{U}}= \left[\overline{\Sigma}_{\set{D}_{k}\set{U}_{n}}\right]_{k=m+1,\ldots,\min(m+B,M),n=1,\ldots,M}$ are obtained using \eqref{acov}. The last equality follows from Definition~\ref{def:gs}.
%
%
$$
\hspace{-1.8mm}
\begin{array}{l}
\displaystyle\overline{\Sigma}_{{\set{U}}\set{D}}
\overline{R}^{-1}_{\set{D}\set{D}}\Sigma_{\set{D}\set{S}}\vspace{0.5mm}\\
\displaystyle =\overline{\Sigma}_{{\set{U}}\set{D}} U^{\top} U \Sigma_{\set{D}\set{S}}\\
\displaystyle =\sum^M_{m=1} \overline{\Sigma}_{{\set{U}}(\set{D}_{m}\cup\set{D}^B_{m})}
\left[\hspace{-1mm}
\begin{array}{c}
U^{\top}_{mm}\\
U^{B\top}_{m}
\end{array}
\hspace{-1mm}\right]\left[U_{mm}, U^B_{m} \right]  \Sigma_{(\set{D}_{m}\cup\set{D}^B_{m})\set{S}}\\
= \displaystyle\sum^M_{m=1} 
\left(U_{mm}\overline{\Sigma}_{{\set{D}_{m}\set{U}}} \hspace{-1mm} +\hspace{-0.5mm}
U^{B}_{m}\overline{\Sigma}_{\set{D}^B_{m}{\set{U}}} \right)^{\hspace{-1mm}\top} \hspace{-1mm}\left(U_{mm}\Sigma_{\set{D}_{m}\set{S}}\hspace{-0.5mm}+\hspace{-0.5mm}
U^B_{m}\Sigma_{\set{D}^B_{m}\set{S}}\right)\\
= \displaystyle\sum^M_{m=1} (\dot{\Sigma}^{m}_{\set{U}})^{\top}\dot{R}_m \dot{\Sigma}^{m}_{\set{S}}\\
= \ddot{\Sigma}_{\set{U}\set{S}}\ .
\end{array}
$$
The fourth equality is due to Lemma~\ref{lemma1} and Definition~\ref{def:ls}. The last equality follows from Definition~\ref{def:gs}.

Finally,
$$
\hspace{-1.8mm}
\begin{array}{l}
\displaystyle{\Sigma}_{\set{S}\set{D}}
\overline{R}^{-1}_{\set{D}\set{D}}\left(y_\set{D}-\mu_\set{D} \right)\vspace{0.5mm}\\
\displaystyle = \Sigma_{\set{S}\set{D}} U^{\top} U \left(y_\set{D}-\mu_\set{D} \right)\\
\displaystyle =\sum^M_{m=1} \Sigma_{\set{S}(\set{D}_{m}\cup\set{D}^B_{m})}
\left[\hspace{-1mm}
\begin{array}{c}
U^{\top}_{mm}\\
U^{B\top}_{m}
\end{array}
\hspace{-1mm}\right] \left[U_{mm}, U^B_{m} \right]\left(y_{\set{D}_{m}\cup\set{D}^B_{m}}-\mu_{\set{D}_{m}\cup\set{D}^B_{m}} \right)\\
= \displaystyle\sum^M_{m=1} 
\left(U_{mm}\Sigma_{\set{D}_{m}\set{S}}+
U^B_{m}\Sigma_{\set{D}^B_{m}\set{S}}\right)^{\hspace{-1mm}\top}\\
\quad\quad\quad\left(U_{mm}\left(y_{\set{D}_m}-\mu_{\set{D}_m} \right)+
U^B_{m} \left(y_{\set{D}^B_{m}}-\mu_{\set{D}^B_{m}} \right)\right)\\
=\displaystyle\sum^M_{m=1}\hspace{-0mm} (\dot{\Sigma}^{m}_{\set{S}})^{\top} \hspace{-0mm}\dot{R}_m \dot{y}_{m}\\
= \ddot{y}_{\set{S}}\ .
\end{array}
$$
The fourth equality is due to Lemma~\ref{lemma1} and Definition~\ref{def:ls}. The last equality follows from Definition~\ref{def:gs}.
$$
\hspace{-1.8mm}
\begin{array}{l} 
{\Sigma}^{\text{\tiny{LMA}}}_{{\set{U}}{\set{U}}}\\
\stackrel{\eqref{vapvar}}{=}\displaystyle\overline{\Sigma}_{\set{U}{\set{U}}}-\overline{\Sigma}_{{\set{U}}\set{D}}\overline{\Sigma}_{\set{D}\set{D}}^{-1}\overline{\Sigma}_{\set{D}{\set{U}}}\\
\stackrel{\eqref{eq:pitck}}{=} \displaystyle{\Sigma}_{\set{U}{\set{U}}} -
\overline{\Sigma}_{{\set{U}}\set{D}}
\overline{R}^{-1}_{\set{D}\set{D}}
\overline{\Sigma}_{\set{D}{\set{U}}} + \overline{\Sigma}_{{\set{U}}\set{D}}
\overline{R}^{-1}_{\set{D}\set{D}}\Sigma_{\set{D}\set{S}}\ddot{\Sigma}_{\set{S}\set{S}}^{-1}\Sigma_{\set{S}\set{D}}\overline{R}^{-1}_{\set{D}\set{D}}
\overline{\Sigma}_{\set{D}{\set{U}}}\\
=\displaystyle{\Sigma}_{\set{U}{\set{U}}} -
\ddot{\Sigma}_{\set{U}\set{U}} +\ddot{\Sigma}_{\set{U}\set{S}}\ddot{\Sigma}_{\set{S}\set{S}}^{-1}\ddot{\Sigma}^{\top}_{\set{U}\set{S}}\ .
\end{array}
$$
The last equality is derived using Definition~\ref{def:gs}, specifically, from the expression of the $\ddot{\Sigma}_{\set{U}\set{S}}$ component above as well as that of the following $\ddot{\Sigma}_{\set{U}\set{U}}$ component:
$$
\hspace{-1.8mm}
\begin{array}{l}
\displaystyle\overline{\Sigma}_{{\set{U}}\set{D}}
\overline{R}^{-1}_{\set{D}\set{D}}\overline{\Sigma}_{\set{D}{\set{U}}}\vspace{0.5mm}\\
\displaystyle =\overline{\Sigma}_{{\set{U}}\set{D}} U^{\top} U \overline{\Sigma}_{\set{D}{\set{U}}}\\
\displaystyle =\sum^M_{m=1} \overline{\Sigma}_{{\set{U}}(\set{D}_{m}\cup\set{D}^B_{m})}
\left[\hspace{-1mm}
\begin{array}{c}
U^{\top}_{mm}\\
U^{B\top}_{m}
\end{array}
\hspace{-1mm}\right]\left[U_{mm}, U^B_{m} \right]  \overline{\Sigma}_{(\set{D}_{m}\cup\set{D}^B_{m}){\set{U}}}\\
= \displaystyle\sum^M_{m=1} 
\left(U_{mm}\overline{\Sigma}_{{\set{D}_{m}\set{U}}} \hspace{-0.5mm} +\hspace{-0.5mm}
U^{B}_{m}\overline{\Sigma}_{\set{D}^B_{m}{\set{U}}} \right)^{\hspace{-1mm}\top}\hspace{-1mm}\left(U_{mm}\overline{\Sigma}_{{\set{D}_{m}\set{U}}} \hspace{-0.5mm} +
\hspace{-0.5mm} U^{B}_{m}\overline{\Sigma}_{\set{D}^B_{m}{\set{U}}} \right)\\
=\displaystyle\sum^M_{m=1}(\dot{\Sigma}^{m}_{\set{U}})^{\top} \dot{R}_m \dot{\Sigma}^{m}_{\set{U}}\\
= \ddot{\Sigma}_{\set{U}\set{U}}\ .
\end{array}
$$
The fourth equality is due to Lemma~\ref{lemma1} and Definition~\ref{def:ls}. The last equality follows from Definition~\ref{def:gs}.
\section{Parallel Computation of $\overline{\Sigma}_{\set{D}_{m}\set{U}}$ and $\overline{\Sigma}_{\set{D}^B_{m}\set{U}}$}
\label{app4}
Computing $\overline{\Sigma}_{\set{D}_{m}\set{U}}$ and $\overline{\Sigma}_{\set{D}^B_{m}\set{U}}$ terms \eqref{acov} requires evaluating $\overline{R}_{\set{D}_{m}\set{U}}$ and $\overline{R}_{\set{D}^B_{m}\set{U}}$ terms \eqref{ares}, which are stored and used by each machine/core $m$ to construct the $m$-th local summary. 
We will describe the parallel computation of $\overline{R}_{\set{D}\set{U}}$, from which $\overline{R}_{\set{D}_{m}\set{U}}$ and $\overline{R}_{\set{D}^B_{m}\set{U}}$ terms can be obtained by machine/core $m$.
To simplify exposition, we will consider the simple setting of $B=1$ and $M=4$ here.

For the blocks \emph{within} the $1$-block band of $\overline{R}_{\set{D}\set{U}}$ (i.e., $|m-n|\leq 1$), they correspond exactly to that of the residual covariance matrix ${R}_{\set{D}\set{U}}$. So, each machine/core $m$ for $m=1,\ldots,4$ can directly compute its respective blocks in parallel (Fig.~\ref{12}), specifically,
$\overline{R}_{\set{D}_{m}\bigcup^{\min(m+1,M)}_{n=\max(m-1,1)}\set{U}_n} = {R}_{\set{D}_{m}\bigcup^{\min(m+1,M)}_{n=\max(m-1,1)}\set{U}_n}$
and
$\overline{R}_{\set{D}^1_{m}\bigcup^{\min(m+2,M)}_{n=\max(m,1)}\set{U}_n} = {R}_{\set{D}_{m+1}\bigcup^{\min(m+2,M)}_{n=\max(m,1)}\set{U}_n}$.
%
\begin{figure}
\includegraphics[scale=0.6]{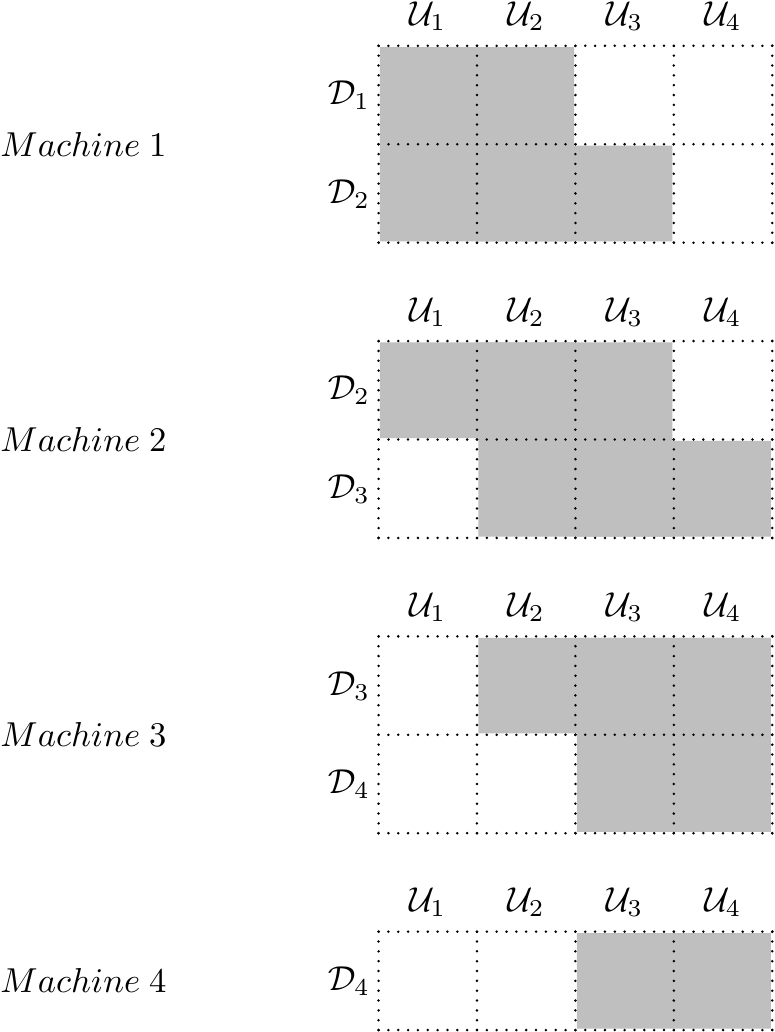}
\caption{Each machine can directly compute its light gray blocks in parallel.}
\label{12}
\end{figure}

Let the upper $N$-diagonal blocks denote the ones that are $N$ blocks above the main diagonal ones of $\overline{R}_{DU}$.
For the blocks \emph{strictly above} the $1$-block band of $\overline{R}_{\set{D}\set{U}}$ (i.e., $n-m>1$), 
the key idea is to exploit the recursive definition of $\overline{R}_{\set{D}\set{U}}$ to compute the upper $2$-diagonal blocks  in parallel using the upper $1$-diagonal blocks, followed by computing the upper $3$-diagonal block using an upper $2$-diagonal block.
Specifically, they can be computed in $2$ recursive steps (Fig.~\ref{13}):
In each recursive step $i$, each machine/core $m$ for $m=1,\ldots,3-i$ uses the upper $i$-diagonal blocks $\overline{R}_{\set{D}_{m+1}\set{U}_{m+1+i}}$ to compute the upper ($i$+$1$)-diagonal blocks $\overline{R}_{\set{D}_{m}\set{U}_{m+1+i}} =R_{\set{D}_{m}\set{D}^1_{m}}R^{-1}_{\set{D}^1_{m}\set{D}^1_{m}}\overline{R}_{\set{D}^1_{m}\set{U}_{m+1+i}}=R_{\set{D}_{m}\set{D}_{m+1}}R^{-1}_{\set{D}_{m+1}\set{D}_{m+1}}\overline{R}_{\set{D}_{m+1}\set{U}_{m+1+i}}$ \eqref{ares} in parallel and then 
communicates $\overline{R}_{\set{D}_m\set{U}_{m+1+i}}$ to machine/core $m-1$. 
\begin{figure}
\includegraphics[scale=0.7]{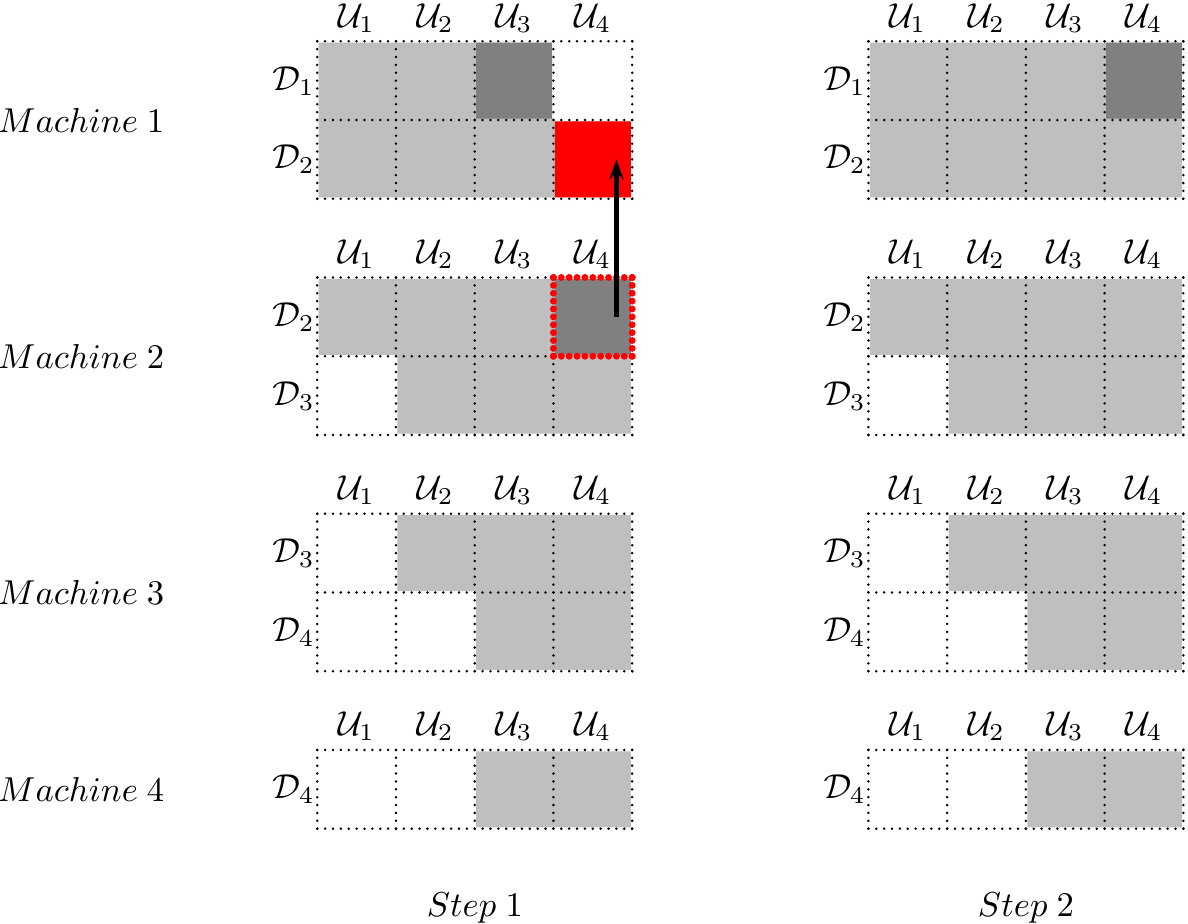}
\caption{In step $1$, machines $1$ and $2$ compute their dark gray blocks in parallel and machine $2$ communicates the dark gray block highlighted in red to machine $1$. In step $2$, machine $1$ computes the dark gray block.}
\label{13}
\end{figure}

For the blocks \emph{strictly below} the $1$-block band of $\overline{R}_{\set{D}\set{U}}$ (i.e., $m-n>1$), it is similar but less straightforward: It can be observed from \eqref{ares} that
$\overline{R}_{\set{D}_m\set{U}_{n}}=\overline{R}_{\set{D}_{m}\set{D}^1_{n}}R_{\set{D}^1_{n}\set{D}^1_{n}}^{-1} R_{\set{D}^1_{n}\set{U}_{n}} = \overline{R}_{\set{D}_{m}\set{D}_{n+1}}R_{\set{D}_{n+1}\set{D}_{n+1}}^{-1} R_{\set{D}_{n+1}\set{U}_{n}}$.
But, machine/core $m$ does not store data associated with the set $D_{n+1}$ of inputs in order to compute $\overline{R}_{\set{D}_{m}\set{D}_{n+1}}$.
To resolve this, the trick is to compute the transpose of $\overline{R}_{\set{D}_m\set{U}_{n}}$ (i.e., $\overline{R}_{\set{U}_{n}\set{D}_m}$) for $m-n>1$ instead. These transposed blocks can also be computed in $2$ recursive steps like the above:
In each recursive step $i$, each machine/core $n$ for $n=1,\ldots,3-i$ uses the upper $i$-diagonal blocks $\overline{R}_{\set{D}_{n+1}\set{D}_{n+1+i}}$ (i.e., equal to ${R}_{\set{D}_{n+1}\set{D}_{n+2}}$ when $i=1$ that can be directly computed by machine/core $n$ in parallel) of $\overline{R}_{\set{D}\set{D}}$ to compute the upper ($i$+$1$)-diagonal blocks $\overline{R}_{\set{U}_{n}\set{D}_{n+1+i}}$ of $\overline{R}_{\set{U}\set{D}}$:
$$
\begin{array}{rl}
\overline{R}_{\set{U}_{n}\set{D}_{n+1+i}} \stackrel{\eqref{ares}}{=}& R_{\set{U}_{n}\set{D}^1_{n}}R^{-1}_{\set{D}^1_{n}\set{D}^1_{n}}\overline{R}_{\set{D}^1_{n}\set{D}_{n+1+i}}\\
=& R_{\set{U}_{n}\set{D}_{n+1}}R^{-1}_{\set{D}_{n+1}\set{D}_{n+1}}\overline{R}_{\set{D}_{n+1}\set{D}_{n+1+i}}
\end{array}
$$
as well as the upper ($i$+$1$)-diagonal blocks $\overline{R}_{\set{D}_{n}\set{D}_{n+1+i}}$ of $\overline{R}_{\set{D}\set{D}}$ in parallel:
$$
\begin{array}{rl}
\overline{R}_{\set{D}_{n}\set{D}_{n+1+i}} \stackrel{\eqref{ares}}{=}& R_{\set{D}_{n}\set{D}^1_{n}}R^{-1}_{\set{D}^1_{n}\set{D}^1_{n}}\overline{R}_{\set{D}^1_{n}\set{D}_{n+1+i}}\\
=& R_{\set{D}_{n}\set{D}_{n+1}}R^{-1}_{\set{D}_{n+1}\set{D}_{n+1}}\overline{R}_{\set{D}_{n+1}\set{D}_{n+1+i}}
\end{array}
$$
and then communicates $\overline{R}_{\set{D}_n\set{D}_{n+1+i}}$ to machine/core $n-1$. 
Finally, each machine/core $n$ for $n=1,2$ transposes the previously computed $\overline{R}_{\set{U}_{n}\set{D}_m}$ back to
$\overline{R}_{\set{D}_m\set{U}_{n}}$ for $m-n>1$ and communicates $\overline{R}_{\set{D}_m\set{U}_{n}}$ to machines/cores $m-1$ and $m$.
%

The parallel computation of $\overline{R}_{\set{D}\set{U}}$ is thus complete. The procedure to parallelize the computation of $\overline{R}_{\set{D}\set{U}}$ for any general setting of $B$ and $M$ is similar to the above, albeit more tedious notationally.

%
\section{Toy Example}
\label{1dtoy}
For our LMA method, the settings are $M=4$, Markov order $B=1$, support set (black $\times$'s) size $|\mathcal{S}|=16$, and training data (red $\times$'s) size $|\mathcal{D}|=400$ such that $x <-2.5$ if $x\in\mathcal{D}_1$, $-2.5\leq x < 0$ if $x\in \mathcal{D}_2$, $0\leq x <2.5$ if $x\in\mathcal{D}_3$, $x \geq 2.5$ if $x\in\mathcal{D}_4$,
and $|\mathcal{D}_1|=|\mathcal{D}_2|=|\mathcal{D}_3|=|\mathcal{D}_4|=100$.
The hyperparameters learned using maximum likelihood estimation are length-scale $\ell=1.2270$, $\sigma_n=0.0939$, $\sigma_s=0.6836$, and $\mu_x=1.1072$. 
It can be observed from Fig.~\ref{fig:1dtoy} that the posterior/predictive mean curve (blue curve) of our LMA method does not exhibit any discontinuity/jump. The area enclosed by the green curves is the $95\%$ confidence region.
\begin{figure}[h]
\includegraphics[scale=0.5]{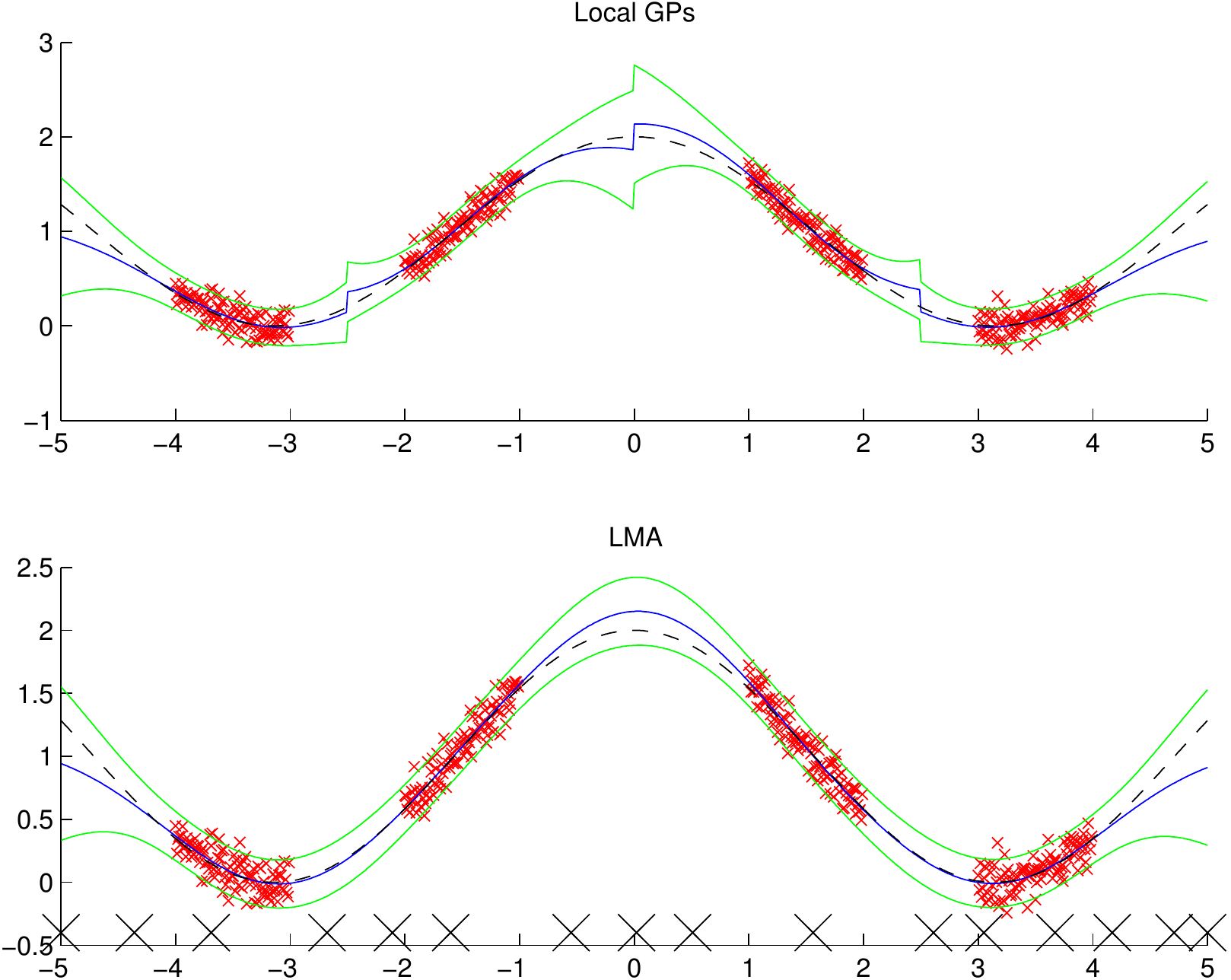}
\caption{The true function (dashed black curve) is $y_x=1+ \text{cos}(x)+0.1\epsilon$ for $-5\leq x\leq 5$ where $\epsilon\sim\mathcal{N}(0,1)$.}
\label{fig:1dtoy}
\end{figure}
%

In contrast, the posterior/predictive mean curve (blue curve) of the local GPs approach 
with $4$ GPs experiences $3$ discontinuities/jumps at the boundaries $x=-2.5,0,2.5$.

\fi
\end{document}